%% file: colm2024_conference.tex
\documentclass{article} %
\usepackage{colm2024_conference}

\usepackage{booktabs}
\usepackage{graphicx}
\usepackage{enumitem}
\usepackage{wrapfig}
\usepackage{algorithm}
\usepackage{algpseudocode}
\usepackage{multicol}

\usepackage{CJKutf8}

\usepackage{microtype}
\usepackage{amsmath}
\usepackage{amsfonts}
\usepackage{colortbl}
\usepackage[utf8]{inputenc}
\usepackage[T1]{fontenc}
\definecolor{lightgray}{rgb}{0.9,0.9,0.9}
\usepackage{caption}
\usepackage{subcaption}
\usepackage{xcolor}
\usepackage{setspace}
\usepackage{url}
\usepackage{multirow}
\usepackage{colortbl}
\usepackage{tabularx}
\usepackage{xltabular}
\usepackage{blindtext}
\usepackage{pgfplots}
\pgfplotsset{compat=1.18} 
\usepackage{tikz}
\usetikzlibrary{er,positioning,bayesnet}
\usepackage{makecell}
\usepackage{tipa}
\usepackage{siunitx}
\usepackage{nicefrac}
\usepackage{tocloft}
\usepackage{listings}
\usepackage[raster,skins]{tcolorbox} %
\usepackage{xltabular}
\usepackage{adjustbox}
\usepackage{xurl}
\usepackage{multirow}
\usepackage{threeparttable}
\usepackage{xcolor}
\input{math_commands.tex}

\title{Qwen3.5-Omni Technical Report}

\author{
\bf Qwen Team
}

\newcommand{\tabincell}[2]{\begin{tabular}{@{}#1@{}}#2\end{tabular}}

\newcommand{\method}{Qwen3.5-Omni\xspace}

\newcommand{\plus}{Qwen3.5-Omni-Plus\xspace}
\newcommand{\flash}{Qwen3.5-Omni-Flash\xspace}

\begin{document}

\maketitle

\begin{abstract}
In this work, we present Qwen3.5-Omni, the latest advancement in the Qwen-Omni model family. Representing a significant evolution over its predecessor, Qwen3.5-Omni scales to hundreds of billions of parameters and supports a 256k context length. By leveraging a massive dataset comprising heterogeneous text-vision pairs and over 100 million hours of audio-visual content, the model demonstrates robust omni-modality capabilities. \plus achieves SOTA results across 215 audio and audio-visual understanding, reasoning, and interaction subtasks and benchmarks, surpassing Gemini-3.1 Pro in key audio tasks and matching it in comprehensive audio-visual understanding.  Architecturally, Qwen3.5-Omni employs a Hybrid Attention Mixture-of-Experts (MoE) framework for both Thinker and Talker, enabling efficient long-sequence inference. The model facilitates sophisticated interaction, supporting over 10 hours of audio understanding and 400 seconds of 720P video (at 1 FPS). To address the inherent instability and unnaturalness in streaming speech synthesis—often caused by encoding efficiency discrepancies between text and speech tokenizers—we introduce ARIA (\textbf{A}daptive \textbf{R}ate \textbf{I}nterleave \textbf{A}lignment). ARIA dynamically aligns text and speech units, significantly enhancing the stability and prosody of conversational speech with minimal latency impact. Furthermore, Qwen3.5-Omni expands linguistic boundaries, supporting multilingual understanding and speech generation across 10 languages with human-like emotional nuance. Beyond preset voices, the model enables zero-shot voice customization via user-provided samples. Finally, Qwen3.5-Omni exhibits superior audio-visual grounding capabilities, generating script-level structured captions with precise temporal synchronization and automated scene segmentation. Remarkably, we observed the emergence of a new capability in omnimodal models: directly performing coding based on audio-visual instructions, which we call Audio-Visual Vibe Coding. Qwen3.5-Omni is publicly accessible via API\footnote{\url{https://www.alibabacloud.com/help/en/model-studio/qwen-omni}}.

\end{abstract}
\begin{figure}[tbh]
    \centering
    \includegraphics[width=\textwidth]{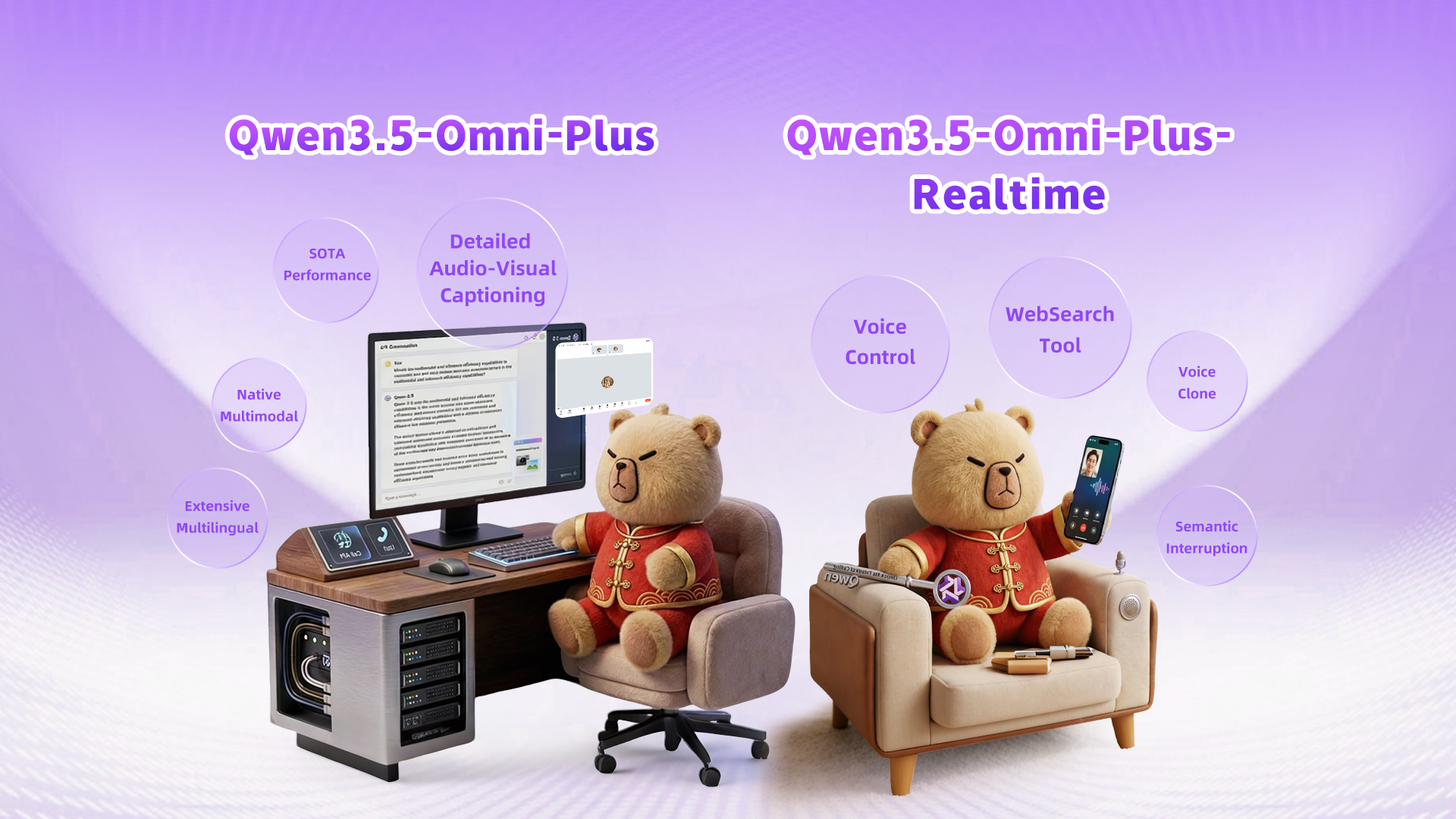}
    \caption{\method~is a unified end-to-end model capable of processing multiple modalities, such as text, audio, image and video, and generating real-time text or speech response. Based on these features, \method~supports a wide range of tasks, including but not limited to voice dialogue, video dialogue, and audio-visual tool use.}
    \label{fig:intro}
\end{figure}

\input{content/intro.tex}
\input{content/arch.tex}

\input{content/pretraining}

\input{content/posttraining}

\input{content/experiments.tex}

\input{content/conclusion.tex}

\clearpage
\bibliography{biblio}
\bibliographystyle{colm2024_conference}

\clearpage
\input{content/authors.tex}

\clearpage
\input{content/appendix.tex}

\end{document}

%% file: math_commands.tex
\usepackage{amsmath,amsfonts,bm}

\def\eqref#1{equation~\ref{#1}}

\def\1{\bm{1}}

\DeclareMathAlphabet{\mathsfit}{\encodingdefault}{\sfdefault}{m}{sl}
\SetMathAlphabet{\mathsfit}{bold}{\encodingdefault}{\sfdefault}{bx}{n}



%% file: content/intro.tex
\section{Introduction}
\label{sec:intro}

Human interaction with the world is inherently omnimodal and agentic, involving the integration of visual, auditory, and linguistic information, and the production of responses through text, speech, and goal-directed tool-mediated actions, facilitating information exchange with other organisms and demonstrating intelligence. Building on the rapid advances in the understanding and reasoning capabilities of large models across text~\citep{gpt3,gpt4,gemini,claude,claude2,claude3,qwen,qwen2,qwen3,llama2,llama3}, vision~\citep{blip2,llava,minigpt-4,qwenvl,qwen2.5vl}, and audio~\citep{qwenaudio, qwen2-audio}, natively omnimodal systems that jointly process and generate across all modalities have drawn substantial attention~\citep{gpt4o,gemini2.5,qwen2.5omni,qwen3omni}. 
However, existing models predominantly operate within passive perception-response paradigms and exhibit limited capacity for scalable agentic behavior, real-time interaction, autonomous tool utilization, and cross-modal reasoning, which are essential prerequisites for practical deployment.

In this report, we present \method, Qwen's latest generation of fully omnimodal LLM, supporting the understanding of text, images, audio, and audio-visual content. Natively pretrained in an omnimodal manner on massive amounts of text, visual data, and more than 100 million hours of audio-visual data, \method is designed as a native omni agent model: it not only perceives and reasons across all modalities, but also acts, autonomously invoking WebSearch, executing complex FunctionCall, generating speech outputs, and engaging in real-time streaming interaction. The model series includes Plus and Flash variants, all of which are instruct models with 256k-token long-context input.

\method builds on the Thinker--Talker architecture introduced in Qwen2.5-Omni~\citep{qwen2.5omni} and introduces \textbf{five key technical upgrades} over Qwen3-Omni~\citep{qwen3omni}: (1) both the Thinker and Talker adopt Hybrid-Attention Mixture-of-Experts (MoE) designs, enabling highly efficient inference; (2) supporting long-context modeling up to 256k tokens, supporting more than 10 hours of audio and over 400 seconds of 720P audio-visual content at 1 FPS;  (3) on the speech generation side, a multi-codebook codec representation enables single-frame, immediate synthesis; (4) the Talker introduces \textbf{ARIA}, a technique that dynamically aligns text and speech units during streaming decoding, significantly improving naturalness and robustness; and (5) multilingual training is substantially expanded, covering 113 languages and dialects for speech recognition and 36 for speech synthesis.

Enabled by these technical advances, \method delivers \textbf{three major new capabilities} over Qwen3-Omni: (1) controllable audio-visual captioning, capable of generating controllable, detailed, and structured captions as well as screenplay-level fine-grained descriptions, including automatic segmentation, timestamp annotation, and detailed descriptions of characters and their relationship to audio; (2) comprehensive real-time interaction, encompassing semantic interruption through native turn-taking intent recognition, end-to-end voice control over volume, speed, and emotion, and voice cloning from user-provided samples; and (3) native omnimodal agentic behavior, including autonomous WebSearch, complex FunctionCall invocation, and Audio-Visual Vibe Coding, an emergent capability wherein the model directly generates executable code from audio-visual instructions, enabling the model to respond to real-time queries without external orchestration.

Critically, \method maintains state-of-the-art performance on text and visual modalities without degradation relative to same-size single-model Qwen counterparts. Across 215 audio and audio-visual understanding, reasoning, and interaction subtasks and benchmarks, covering audio-visual benchmarks, audio benchmarks, ASR benchmarks, language-specific speech-to-text translation tasks, and language-specific ASR tasks, \method-Plus achieves SOTA results, surpassing Gemini-3.1 Pro across general audio understanding, reasoning, recognition, translation, and dialogue, while its overall audio-visual understanding reaches the level of Gemini-3.1 Pro.

%% file: content/arch.tex
\section{Architecture}
\begin{figure}[tbh]
    \centering
    \includegraphics[width=0.9\textwidth]{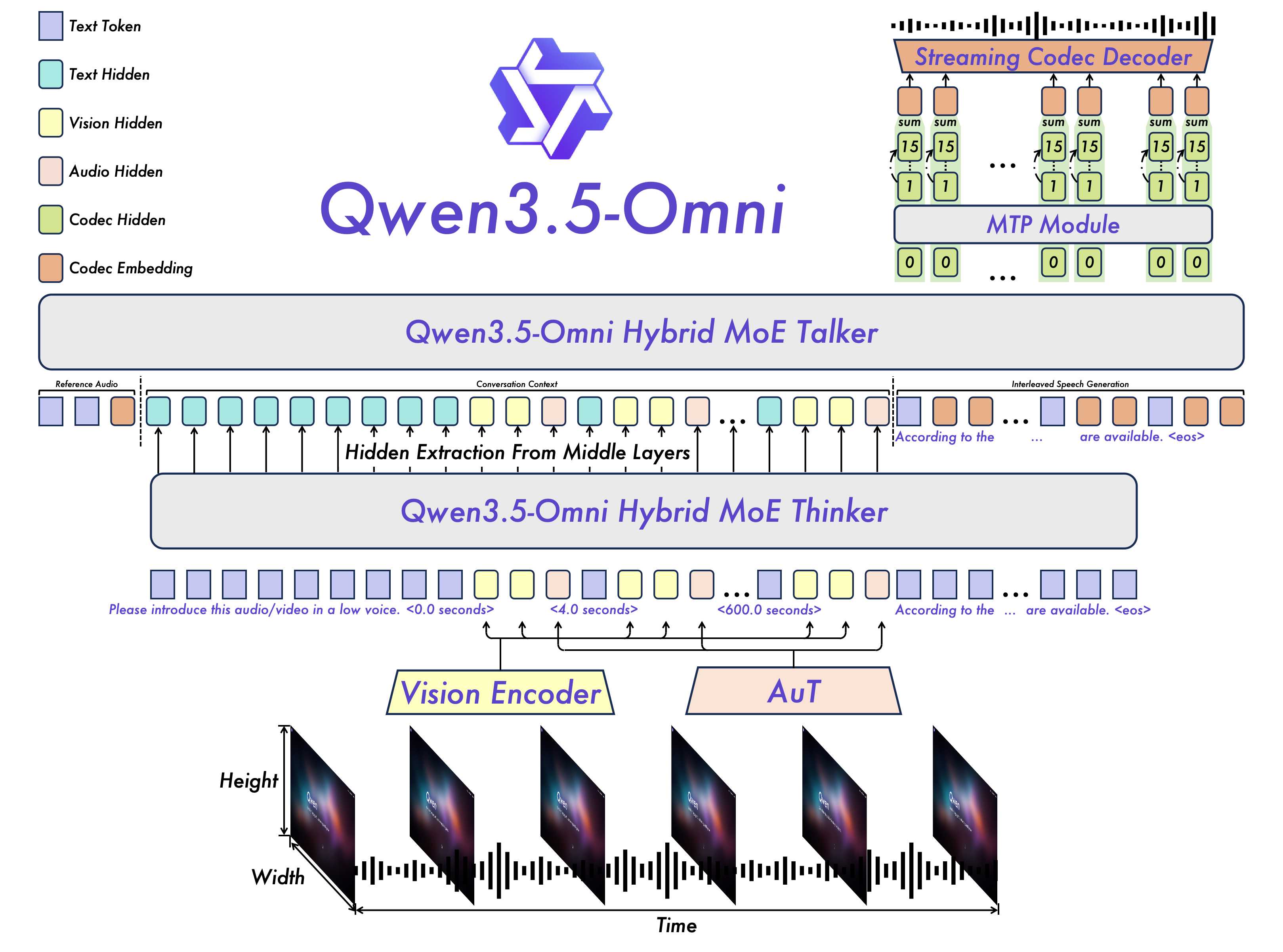}
    \caption{The overview of \method. \method adopts the Thinker-Talker architecture. Thinker is tasked with text generation while Talker focuses on generating streaming speech tokens by receives high-level representations directly from Thinker. To achieve ultra–low-latency streaming, Talker autoregressively predicts a multi-codebook sequence. At each decoding step, an MTP module outputs the residual codebooks for the current frame, after which the Code2Wav renderer incrementally synthesizes the corresponding waveform, enabling frame-by-frame streaming generation.}
    \label{fig:ovewview_arch}
\end{figure}

\subsection{Overview}
As shown in Figure~\ref{fig:ovewview_arch}, \method continues to adopt the Thinker-Talker architecture~\citep{qwen2.5omni}. Compared with Qwen3-Omni~\citep{qwen3omni}, \method introduces several key improvements in scalability, alignment, and real-time interaction:
\begin{itemize}
\item The overall backbone adopts a Hybrid Mixture-of-Experts (MoE) design, improving scalability while better balancing capacity and efficiency across multimodal understanding and generation.
\item The Thinker receives visual and audio signals through the Vision Encoder and AuT, respectively. Audio and video inputs are interleaved for unified multimodal modeling, with explicit timestamps inserted to improve temporal perception, especially for long video or audio-video contexts. This design enables the Thinker to handle extended inputs, supporting up to 256k tokens, 10 hours of audio, or 400 seconds of 720P video at 1 FPS.
\item The Talker is responsible for contextual speech generation by conditioning on multimodal inputs together with the textual outputs from the Thinker. \method adopts the RVQ-based speech representation introduced in Qwen3-Omni~\citep{qwen3omni}, which substantially improves inference efficiency.
\item To support real-time interaction, \method adopts both chunk-wise streaming input processing in the Thinker and a streaming Talker design, enabling low-latency end-to-end multimodal conversation.
\item Different from the dual-track Talker input design in Qwen3-Omni~\citep{qwen3omni}, the Talker in \method adopts ARIA to dynamically align text and speech units before interleaving them. This design mitigates the instability caused by mismatched tokenization rates between text and speech, thereby reducing issues such as skipped words, incorrect pronunciations, and ambiguous rendering of numbers.
\end{itemize}

In the following sections, we first introduce with the AuT encoder, including its training methodology. Then, describe how Thinker processes various inputs. We then detail Talker’s multi-codebook streaming speech generation. Finally, we highlight a series of improvements on both the understanding and generation modules aimed at achieving ultra–low-latency, end-to-end streaming audio inference.

\subsection{Audio Transformer~(AuT)}

\begin{figure}[tbh]
    \centering
    \includegraphics[width=0.6\textwidth]{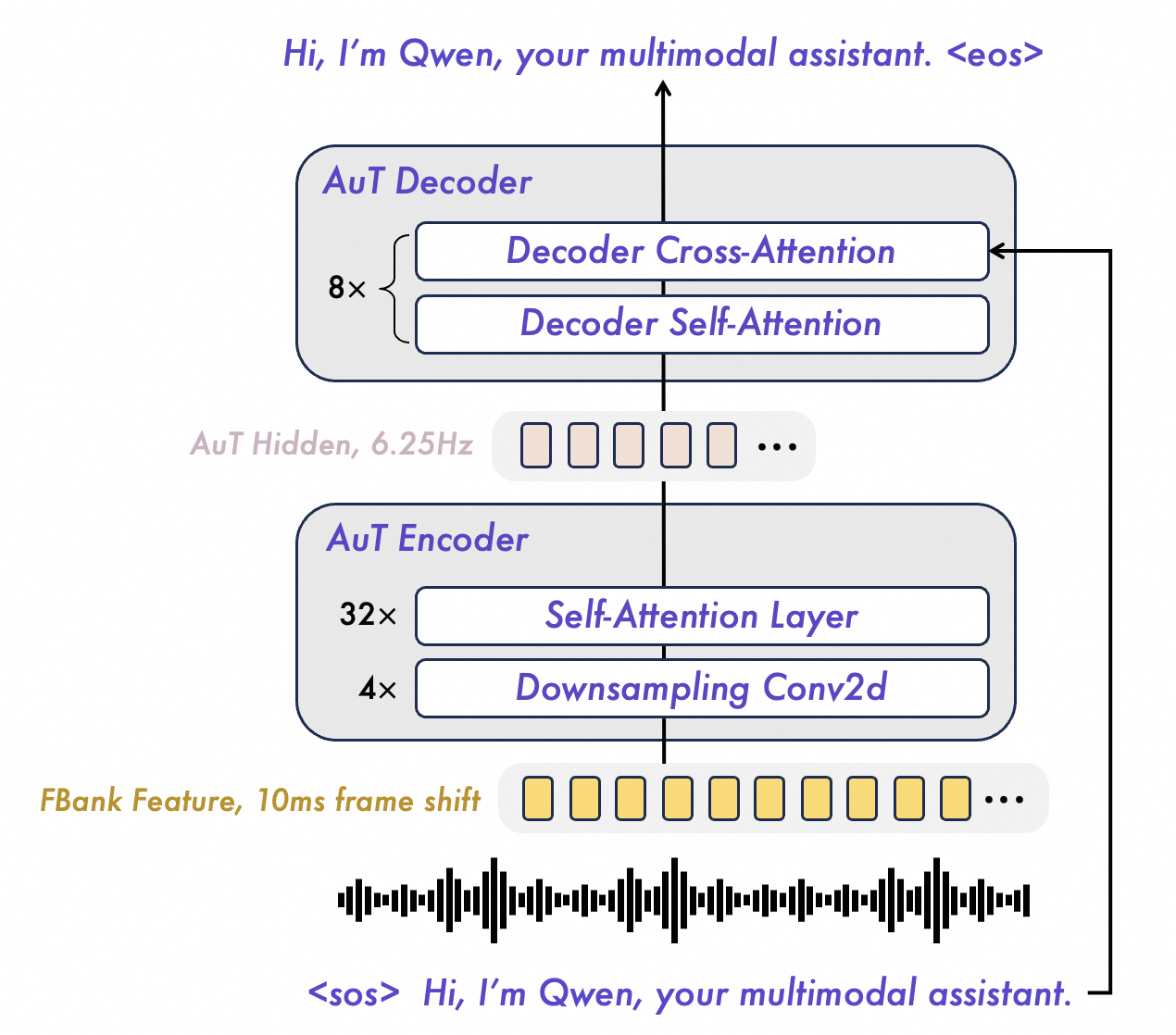}
    \caption{The overview of AuT. Consuming 40 million hours of supervised data especially more multilingual data, AuT encoder in \method obtain stronger general purpose audio representation in 6.25Hz.}
    \label{fig:aut}
\end{figure}

We use transformer based audio encoder trained from scratch in attention-encoder-decoder model AuT, as is shown in Figure~\ref{fig:aut}. The training of \method encoder consumed 40 million hours of audio-text pair data generated by Qwen3-ASR. The filter bank features of the audio are downsampled 16 times using 4 Conv2D blocks and then fed into self-attention layers to obtain audio tokens in 6.25Hz token rate. Comparing to the training process of Qwen3-Omni encoder, the encoder of \method adapts more multilingual data of more than 20 languages, and the proportion of Chinese, English and multilingual data comes to 3.5 : 3.5 : 3. The dynamic attention window size training mechanism is adopted for guaranting balance performance of inference under real-time prefill caching and for the offline audio understanding tasks.

\subsection{Perceivation}
\paragraph{Text, Audio, Image and Video (w/o Audio).} 

The Thinker converts text, audio, image, and silent video inputs into a unified sequence of representations. For text, we use the Qwen3.5 tokenizer~\citep{qwen35blog}, which adopts byte-level byte-pair encoding with a vocabulary size of 250k (up from 150k), improving encoding and decoding efficiency by 10--60\% across most languages. For audio inputs, including audio extracted from video, we resample the waveform to 16 kHz and convert it into a 128-channel mel-spectrogram using a 25 ms window and a 10 ms hop size. We use AuT as the audio encoder, trained from scratch on 40 million hours of audio data, where each output frame corresponds to approximately 160 ms of the original signal. For visual inputs, we adopt the vision encoder from Qwen3.5~\citep{qwen35blog} to process both images and videos. Trained on a mixture of image and video data, this encoder provides strong capabilities in both image understanding and video comprehension. To preserve video information as much as possible while maintaining alignment with the audio stream, we sample video frames at a dynamic frame rate.

\paragraph{Audio-visual Timestamp.}

Following Qwen3-Omni~\citep{qwen3omni}, we apply TM-RoPE to endow the model with temporal awareness for audio-video synchronization. However, we find that directly encoding absolute time through temporal position IDs can lead to excessively sparse indices for visual patches from long video with audio inputs, which weakens long-range temporal modeling. In addition, such a design often requires large-scale and uniformly distributed training samples across different frame rates, increasing data construction cost. To address these issues, we prepend each video or audio-video temporal patch with an explicit timestamp represented as a formatted text string in seconds, allowing the model to learn timecode representations more naturally. For audio sequences, we further insert timestamps at random intervals to improve temporal alignment across modalities. Although this strategy slightly increases the context length, it enables more precise and robust temporal perception, especially when extrapolating long-context multimodal inputs.

In the context of multimodal audio-visual streams, the audio component is encoded with a temporal ID for every 160 ms. The video is treated as a sequence of frames with monotonically increasing temporal IDs that are dynamically adjusted based on their actual timestamps to ensure a consistent temporal resolution of 160 ms per ID. The height and width IDs for video frames are assigned in the same manner as for still images. To prevent positional conflicts when processing multiple modalities, the position numbering is made contiguous, with each subsequent modality commencing from one plus the maximum position ID of the preceding modality. This refined approach to positional encoding enables the model to effectively integrate and jointly model information from diverse modalities. \method aligns these representations using their temporal IDs, which are explicitly anchored to absolute time. This design choice affords the model the flexibility to support streaming inputs of arbitrary duration.

\subsection{Speech Generation}

Talker operates directly on the RVQ tokens produced by \textit{\method-Audio-Tokenizer}. To model the residual codebooks, it employs a multi-token prediction (MTP) module, which enables fine-grained modeling and control of acoustic details. Coupled with a causal ConvNet for waveform reconstruction, Talker delivers high-fidelity speech synthesis with low inference latency and modest computational overhead.

In multi-turn spoken dialogue, Talker is conditioned on the rich contextual information provided by the Thinker component, including historical text tokens, multimodal representations, and the streamed text of the current turn. Such conditioning allows Talker to dynamically modulate acoustic attributes—such as prosody, loudness, and emotion—in accordance with the evolving conversational context.

Architecturally, our approach differs from Qwen3-Omni~\citep{qwen3omni} in two key respects. First, we introduce a dedicated system prompt for Talker that specifies target voice characteristics, thereby enabling both zero-shot voice cloning and controllable speech generation. Compared with conventional speaker embeddings, this prompt can encode richer multimodal cues, including textual descriptions and codec sequences, providing substantially finer-grained control over acoustic realization. Second, we propose \textbf{ARIA} (\textbf{A}daptive \textbf{R}ate \textbf{I}nterleave \textbf{A}lignment), which unifies the conventional dual-channel generation paradigm into a single-channel formulation. Rather than relying on MFA-derived alignments or fixed interleaving rates, ARIA enforces an adaptive rate constraint: for any prefix of the generated sequence, the cumulative speech-to-text token ratio must not exceed the corresponding item-level global ratio. Despite its simplicity, this design affords flexible text-speech alignment across languages, including those with relatively low encoding efficiency, and naturally supports arbitrary text-token prefixes followed by coherent speech-token continuation.

\subsection{Designs for Streaming and Concurrency}
In streaming audio-visual interaction scenarios, the first-packet latency is a critical factor affecting user experience, and the model's concurrency capability is key to reducing service costs and improving response speed. This section discusses how \method enhances concurrency and reduces first-packet latency through algorithmic and architectural optimizations. Table~\ref{tab:inference-overview} provides an overview of the relevant architecture of the \method and its associated latency.

\begin{table}[htbp]
\centering
\caption{\textbf{Architecture of \method and end-to-end first-packet latency under audio/video settings (ms).}}
\label{tab:inference-overview}
\vspace{-1mm}
\begin{tabular}{lcc}
\toprule
\textbf{Module} & \textbf{Architecture} & \textbf{Streaming} \\
\midrule
Audio Encoder  & AuT                     & $\checkmark$ \\
Vision Encoder & SigLIP2                 & -- \\
Thinker        & Hybrid MoE Transformer  & $\checkmark$ \\
Talker         & Hybrid MoE Transformer  & $\checkmark$ \\
MTP            & Dense Transformer       & $\checkmark$ \\
Code2wav       & ConvNet                 & $\checkmark$ \\
\midrule
\multicolumn{2}{l}{\textbf{First-Packet Latency (Audio Input)}} 
& \textit{Plus:} 435ms \quad \textit{Flash:} 235ms \\
\multicolumn{2}{l}{\textbf{First-Packet Latency (Video Input)}} 
& \textit{Plus:} 651ms \quad \textit{Flash:} 426ms \\
\bottomrule
\end{tabular}
\end{table}

\paragraph{Chunked Prefilling and Hybrid MoE Architecture.} In \method, we retain the chunked-prefilling mechanism as implemented in Qwen3-Omni and Qwen2.5-Omni, whose audio and vision encoders are capable of outputting chunks along the temporal dimension. This approach significantly reduces the Time-To-First-Token (TTFT) for both the Thinker and the Talker. Architecturally, both the Thinker and the Talker in \method are built upon the Hybrid MoE architecture introduced in Qwen3.5. Beyond the general efficiency advantage of Hybrid MoE, this architecture includes the Gated Delta Net (GDN) module, which is particularly effective for accelerating the modeling of long audio-video sequences. As a result, it significantly reduces KV-cache I/O overhead in long-context inference, improving generation throughput and enabling higher serving concurrency.

\paragraph{Streaming Generation with ARIA.} For streaming speech generation and high-concurrency serving, \method largely inherits the efficient design of Qwen3-Omni: Talker predicts RVQ codec tokens with a lightweight MTP module, and the generated multi-codebook tokens are converted to waveform by a causal and streaming ConvNet codec decoder. These components remain computationally lightweight, batch-friendly, and well-suited for low-latency deployment. Built on this shared foundation, the previously introduced ARIA further reformulates the dual-channel generation pattern in Qwen3-Omni into a unified interleaved single-stream formulation over text and speech tokens. By organizing text and speech generation under a monotonic interleaving constraint, ARIA reduces the synchronization overhead between separate generation tracks, enables more efficient token scheduling during decoding, and better matches the naturally incremental regime of streaming interaction.

In Table~\ref{tab:inference-lantency}, we report the theoretical first-packet latency of \method under different concurrency levels for audio and video input, evaluated on internal vLLM with \textit{torch.compile} and CUDA Graph acceleration enabled for the MTP module and codec decoder. Here, Thinker TTFT (\textit{Time-To-First-Token}) denotes the time from receiving the input stream to the first text token generated by Thinker, while Talker TTFC (\textit{Time-To-First-Chunk}) measures the time until Talker produces the first audio chunk. TPOP (\textit{Time-Per-Output-Token}) represents the per-output-token latency during steady-state decoding, where Talker TPOP includes the combined latency of the Talker backbone and the MTP module. TPS (\textit{Tokens Per Second}) denotes generation throughput. Since ARIA organizes text and speech generation in a unified interleaved stream, Overall Latency cannot be obtained by simply summing several row values, but instead reflects the end-to-end critical path to the first playable audio packet. We also note that, due to the substantial scale difference between \method-Flash and \method-Plus, the two variants adopt different deployment-time resource allocation and parallelization strategies; therefore, their latency and throughput numbers are not intended for strict horizontal comparison. As shown in the table, \method maintains stable latency and decoding efficiency as concurrency increases, while the low Generation RTF provides sufficient margin for smooth streaming audio generation.

\begin{table}[htbp]
\centering
\caption{\textbf{Theoretical first-packet latency of \method under different concurrency levels. A/V denotes audio/video input.}}
\label{tab:inference-lantency}
\small
\begin{tabular}{@{}lcccccc@{}}
\toprule
 & \multicolumn{3}{c}{\textbf{\method-Flash}} & \multicolumn{3}{c}{\textbf{\method-Plus}} \\
\cmidrule(lr){2-4} \cmidrule(lr){5-7}
 & \textbf{1 Conc.} & \textbf{4 Conc.} & \textbf{8 Conc.} & \textbf{1 Conc.} & \textbf{4 Conc.} & \textbf{8 Conc.} \\
\midrule
Thinker TTFT & 80/255ms & 86/446ms & 103/765ms & 162/377ms & 183/907ms & 260/1243ms \\
Talker TTFC & 56/61ms & 68/108ms & 81/116ms & 54/56ms & 72/88ms & 95/116ms \\
Thinker TPOP & 5.6/5.9ms & 8.2/9.2ms & 9.6/15.8ms & 17.4/18.5ms & 25.6/26.9ms & 33.3/40.2ms \\
Talker TPOP & 14.2/14.2ms & 16.9/17.0ms & 20.5/20.6ms & 14.9/14.9ms & 21.0/21.3ms & 25.8/27.1ms \\
Codec Decode & \multicolumn{6}{c}{3\textasciitilde5ms} \\
\midrule
\textbf{Overall Latency} & \textbf{235/426ms} & \textbf{298/891ms} & \textbf{352/1625ms} & \textbf{435/651ms} & \textbf{619/1515ms} & \textbf{955/1980ms} \\
\midrule
Thinker TPS & 177/171 & 556/457 & 942/598 & 57/54 & 156/149 & 266/240 \\
Talker TPS & 70/70 & 237/235 & 389/388 & 67/67 & 191/189 & 320/296 \\
\midrule
\textbf{Generation RTF} & \textbf{0.178} & \textbf{0.211} & \textbf{0.257} & \textbf{0.187} & \textbf{0.267} & \textbf{0.334} \\
\bottomrule
\end{tabular}
\end{table}

%% file: content/pretraining.tex
\section{Pretraining}
\begin{table}[htbp]
\centering
\caption{\textbf{Supported languages and dialects in \method-Plus.}}
\vspace{-1mm}
\begin{tabular}{lcp{11cm}}
\toprule
Modality & \# Varieties & Supported languages and dialects \\
\midrule
Text & 201 & See Qwen3.5 for the complete list of supported languages. \\
Speech Input & 113 & \textbf{74 languages:} Afrikaans, Arabic, Asturian, Azerbaijani, Basque, Belarusian, Bengali, Bosnian, Bulgarian, Cantonese, Catalan, Cebuano, Chinese, Croatian, Czech, Danish, Dutch, English, Esperanto, Estonian, Filipino, Finnish, French, Galician, Georgian, German, Greek, Hebrew, Hindi, Hungarian, Icelandic, Indonesian, Interlingua, Italian, Japanese, Javanese, Kannada, Kazakh, Korean, Kyrgyz, Lingala, Latvian, Lithuanian, Macedonian, Malay, Malayalam, Maltese, Maori, Marathi, Mongolian, Norwegian Bokmål, Norwegian Nynorsk, Oriya, Persian, Polish, Portuguese, Punjabi, Romanian, Russian, Serbian, Slovak, Slovenian, Spanish, Swahili, Swedish, Tajiki, Tamil, Telugu, Thai, Turkish, Ukrainian, Urdu, Uyghur, and Vietnamese.

\textbf{39 Chinese dialects:} Northeastern Mandarin, Guizhou dialect, Guangdong Cantonese, Henan dialect, Hong Kong Cantonese, Shanghainese, Shaanxi dialect, Tianjin dialect, Taiwanese Mandarin, Yunnan dialect, Anhui dialect, Fujian dialect, Gansu dialect, Guangdong Mandarin, Hubei dialect, Hunan dialect, Jiangxi dialect, Shandong dialect, Shanxi dialect, Sichuanese, Guangxi dialect, Hainan dialect, Chongqing dialect, Changsha dialect, Hangzhou dialect, Hefei dialect, Yinchuan dialect, Zhengzhou dialect, Shenyang dialect, Wenzhou dialect, Wuhan dialect, Kunming dialect, Taiyuan dialect, Nanchang dialect, Jinan dialect, Lanzhou dialect, Nanjing dialect, Hakka, and Southern Min. \\
Speech Output & 36 & \textbf{29 languages:} Chinese, English, German, Italian, Portuguese, Spanish, Japanese, Korean, French, Russian, Thai, Indonesian, Arabic, Vietnamese, Turkish, Finnish, Polish, Hindi, Dutch, Czech, Urdu, Tagalog, Swedish, Danish, Hebrew, Icelandic, Malay, Norwegian, and Persian.

\textbf{7 Chinese dialects:} Sichuanese, Beijing dialect, Tianjin dialect, Nanjing dialect, Shaanxi dialect, Cantonese, and Southern Min. \\
\bottomrule
\end{tabular}
\label{table:languages}
\end{table}

\method is pre-trained on a diverse dataset that encompasses multiple languages and dialects as shown in Table~\ref{table:languages} and modalities, including image-text, video-text, audio-text, video-audio, video-audio-text, and pure text corpora. Following Qwen3-Omni~\citep{qwen3omni}, we employ a wider range of natural language prompts to enhance both the generalization ability and instruction-following capabilities. To achieve robust performance across all modalities, our training strategy incorporates both unimodal and cross-modal data from the early pretraining stage.

In Qwen3-Omni~\citep{qwen3omni}, we employ TMRoPE to endow the model with temporal awareness. However, we identify two key limitations of this approach: (1) By directly tying temporal position IDs to absolute time, it produces excessively large and sparse temporal position IDs for long audio-video or video inputs, which undermines the model’s ability to capture long-range temporal contexts. (2) Effective learning under this scheme typically requires large-scale and uniformly distributed sampling across different frame rates (fps), significantly increasing the cost of training data construction. To address these issues, we prepend each video or audio-video temporal patch with a timestamp represented as a formatted text string in seconds, enabling the model to better learn and interpret timecode representations. In addition, for audio sequences, we insert timestamps at random intervals to better align training across different modalities. Although this approach introduces a modest increase in context length, it allows the model to perceive temporal information more effectively and precisely.

The pre-training of \method is structured into three distinct stages. In the first stage, we lock the LLM parameters and focus on training the vision and audio encoders, utilizing a vast corpus of audio-text and image-text pairs to enhance semantic understanding within the LLM. In the second stage, we unfreeze all parameters and train with a wider range of multimodal data for more comprehensive learning with a sequence length of 32,768. In the final stage, we use data with a sequence length of 262,144 to enhance the model's ability to understand complex long-sequence data:

\begin{enumerate}[label=(\arabic*)]
    \item \textbf{Encoder Alignment Stage (S1)}: During the initial pretraining phase, the LLM component of \method is initialized with parameters from Qwen3.5, while the vision encoder is adopted from Qwen3.5, and the audio encoder is initialized with AuT. The two encoders are trained separately on the fixed LLM, with both initially focusing on training their respective adapters before training the encoders.
    \item \textbf{General Stage (S2)}: The second phase of pretraining utilizes a large-scale dataset containing approximately 4 trillion tokens, with the following distribution across modalities: text (0.92 trillion), audio (1.99 trillion), image (0.95 trillion), video (0.14 trillion), and video-audio 0.29 trillion). During this stage, the introduction of more diverse multimodal data and tasks enhances the model’s understanding and interaction capabilities in auditory, visual, textual, and audio-visual information.
    \item \textbf{Long Context Stage (S3)}: In the final pre-training phase, we increased the maximum token length from 32,768 to 262,144 and also raised the proportion of long audio and long video in the training data. Experimental results indicate that these adjustments lead to significant improvements in the model's ability to understand long sequence data.
\end{enumerate}

%% file: content/posttraining.tex
\section{Post-training}\label{sec:post}
\subsection{Thinker}

The post-training phase employs a three-stage strategy for the Thinker, designed to preserve the model's capabilities across all modalities without degradation, ensure high response quality under audio queries, and optimize the overall interaction experience. The training corpus, structured in the ChatML~\citep{chatml} format, encompasses pure text, visual, audio, and mixed-modality conversational data. Specifically, the process consists of the following stages:

\begin{itemize}
    \item \textbf{Stage 1: Specialist Distillation} 
To establish a strong foundation for omnimodal capabilities, we first train a suite of domain-specialized teacher models via independent Supervised Fine-Tuning (SFT) and reinforcement learning (RL). All teacher models are fine-tuned from the pre-trained Qwen-3.5 base checkpoint. Beyond text-related tasks, including agentic, coding, and foundational reasoning tasks, we also train specialized teacher models for vision and audio. These teacher models are used to generate domain-specific data, enabling the specialized capabilities learned in each domain to be distilled into a single unified model.

    \item \textbf{Stage 2: On-Policy Distillation} 
Through the specialist distillation described above, the model already achieves strong performance in domains such as multimodal understanding and reasoning, as well as text-based dialogue, reasoning, coding, and agentic tasks. Nevertheless, a substantial gap remains between the quality of responses conditioned on audio queries and that of responses conditioned on text queries, particularly in speech dialogue. To reduce this gap, we introduce a second-stage training procedure based on on-policy distillation (OPD), with the goal of distilling the model’s stronger response capabilities under text inputs into the audio-input setting. Concretely, for each audio-text paired query, we first obtain a response generated under the text condition, which typically exhibits higher quality in terms of fluency, reasoning, and task completion. We then use this response as the distillation target for the corresponding audio-conditioned query. By training on such on-policy targets, the model gradually aligns its audio-conditioned outputs with its text-conditioned behavior, thereby improving response quality under audio inputs and promoting modality-consistent generation.

    \item \textbf{Stage 3: Interaction-Aligned Reinforcement Learning} 
Although the previous two stages substantially improve the model’s domain capabilities and cross-modal response quality, they are not sufficient to fully optimize the model for real-world interactive use. In multi-turn conversations, we observe several interaction-specific issues, including unintended language code-switching, persona inconsistency, and degraded instruction-following over extended contexts. To mitigate these issues, we introduce Interaction-Aligned RL, a third-stage reinforcement learning procedure aimed at optimizing the model for interaction quality. We construct multi-turn interaction trajectories and design reward signals around these user experience objectives, enabling the model to learn behaviors that are more stable, consistent, and aligned in prolonged interactions. By explicitly optimizing for interaction quality, this stage improves the model’s overall usability in practical conversational scenarios.

\end{itemize}

\subsection{Talker}

We employ a four-stage training pipeline for Talker, enabling \method to generate natural and contextually appropriate spoken responses jointly with text. All training data is organized in the ChatML format to maintain consistency with Thinker and to facilitate voice cloning.

\begin{enumerate}[label=(\arabic*)]
    \item \textbf{General Stage}: In the initial pre-training stage, we train \method on more than 20 million hours of multilingual speech data paired with multimodal context. In particular, the introduction of more diverse tasks, such as instruction-following speech generation, substantially enhance contextual reasoning and paralinguistic alignment, going beyond a simple monotonic mapping from multimodal representations to speech.
    
    \item \textbf{Long-Context Stage}: We perform data quality stratification through a dedicated curation pipeline and conduct continual pre-training (CPT) on high-quality subsets. Augmented by Qwen3-Omni-Captioner, this stage mitigates hallucinations introduced by noisy data in the initial pre-training phase and substantially improves the naturalness and quality of generated speech. Meanwhile, we extend the maximum context length to 64k tokens, allowing the model to better handle long and complex user inputs and to produce more contextually grounded speech responses.
    
    \item \textbf{Reinforcement Learning Stage}: We further align model behavior with human preferences through Direct Preference Optimization (DPO)~\citep{rafailov2024direct}. Concretely, we construct multilingual preference pairs based on human annotations and optimize the model with DPO. In addition, we incorporate rule-based rewards and adopt GSPO~\citep{gspo} to further improve overall capability and training stability across diverse tasks.
    
    \item \textbf{Speaker Fine-tuning Stage}: Finally, we perform lightweight speaker fine-tuning on top of the base model, enabling \method to faithfully capture target speaker characteristics while further improving the naturalness, expressiveness, and controllability of its speech outputs.
\end{enumerate}

%% file: content/experiments.tex
\section{Evaluation}
\label{sec:experiment} 
A comprehensive evaluation was performed on two variants of models, including \method-Flash and \method-Plus. The evaluation results are divided into two main categories: understanding~(X$\to$Text) and speech generation~(X$\to$Speech). 

\subsection{Evaluation of X$\to$Text}
In this section, we evaluate \method's ability to comprehend various multimodal inputs (text, audio, vision, and audio-visual video) and generate textual responses.
\paragraph{Text$\to$Text} Our evaluation of \method on text $\to$ text primarily focuses on general knowledge tasks, instruction following, long context tasks, STEM tasks, reasoning tasks and general agent ability.
Specifically, we utilize
MMLU-Pro \citep{mmlupro}, MMLU-Redux \citep{mmluredux}, SuperGPQA \citep{supergpqa} and C-Eval \citep{ceval} for general knowledge tasks,
IFEval \citep{ifeval} and IFBench \citep{ifbench} for instruction following,
AA-LCR \citep{aalcr} and LongBench v2 \citep{longbenchv2} for long context tasks,
GPQA \citep{gpqa} for STEM tasks,
LiveCodeBench v6 \citep{livecodebench}, HMMT Nov 25 \citep{hmmtnov25} and IMOAnswerBench \citep{imoanswerbench} for reasoning tasks,
BFCL-V4 \citep{bfcl} and TAU2Bench \citep{barres2025tau2} for general agent ability.

\paragraph{Audio$\to$Text} 

To evaluate audio-to-text capabilities, we employ benchmarks across four domains: audio understanding, end-to-end speech dialogue, speech-to-text translation (S2TT), and automatic speech recognition (ASR). For audio understanding, we utilize MMAU~\citep{sakshi2024mmaumassivemultitaskaudio}, MMAR~\citep{mmar}, MMSU~\citep{mmsu}, RUL-MuchoMusic~\citep{zang2025you}, and SongFormBench~\citep{hao2025songformer} to assess comprehension of sound effects, speech, and music. Dialogue performance is evaluated via VoiceBench~\citep{chen2024voicebench}, URO-Bench-pro~\citep{yan2025uro}, SpeechRole~\citep{jiang2025speechrole}, and WildSpeech-Bench~\citep{zhang2025wildspeechbenchbenchmarkingendtoendspeechllms}. For S2TT, we focus on the translation of the top 59 languages in Fleurs~\citep{Conneau2022FLEURSFL} into English and Chinese. Finally, ASR performance is measured using Fleurs~\citep{Conneau2022FLEURSFL}, Common Voice~\citep{DBLP:conf/lrec/ArdilaBDKMHMSTW20}, LibriSpeech~\citep{Librispeech}, WenetSpeech~\citep{DBLP:conf/icassp/ZhangLGSYXXBCZW22}, KeSpeech~\citep{DBLP:conf/nips/Tang0XSLZWTXZYL21}, Opencpop-test~\citep{DBLP:conf/interspeech/WangWZWLXZXB22}, and MIR-1K (vocal)~\citep{DBLP:journals/taslp/HsuJ10}, covering multilingual speech, Chinese dialects, and singing voice transcription.

\paragraph{Vision$\to$Text}
The evaluation of the model's vision-to-text capabilities encompasses a suite of benchmarks targeting diverse and challenging tasks. To assess performance in specialized domain of mathematical and STEM reasoning, we utilize MMMU~\citep{yue2023mmmu}, MMMU-Pro~\citep{mmmupro}, MathVista~\citep{mathvista}, MathVision~\citep{mathvision}, DynaMath~\citep{dynamath}, ZEROBench~\citep{zerobench}. For the general visual question answering, the model is evaluated on RealWorldQA~\citep{mme-realworld}, MMStar~\citep{chen2024we}, HallusionBench~\citep{hallusionbench}, and SimpleVQA~\citep{simplevqa}. The model's proficiency in document understanding is measured using the  CharXiv~\citep{wang2024charxiv}, CC-OCR~\citep{ccocr},  AI2D~\citep{kembhavi2016diagram}, MMLongBench-Doc~\citep{mmlongbench}, and OCRBench~\citep{Liu_2024_OCRBench}. Furthermore, the model's spatial intelligence is specifically tested on ERQA~\citep{erqa}, CountBench~\citep{countbench}, RefCOCO~\citep{refcoco}, ODInW13~\citep{odinw}, and EmbSpatialBench~\citep{embspatial}. To evaluate performance on dynamic visual data, we report results on six video understanding benchmarks: Video-MME~\citep{fu2024video}, MLVU~\citep{mlvu}, MVBench~\citep{li2024mvbench}, LVBench~\citep{lvbench}, MMVU~\citep{mmvu} and MME-VideoOCR~\citep{videoocr}. Specifically, we evaluate the model’s performance on medical VQA across three established benchmarks: SLAKE~\citep{slake}, PMC-VQA~\citep{pmc}, and MedXpertQA-MM~\citep{medxpertqa}. This assessment is designed to demonstrate the model’s comprehensive clinical reasoning capabilities and its potential utility as a reliable healthcare AI assistant.

\paragraph{Audio-Visual Video$\to$Text}
We evaluate our model's audio-visual understanding capabilities from multiple perspectives. For text-query evaluation, we use DailyOmni~\citep{dailyomni}, WorldSense~\citep{worldsense}, AVUT~\citep{avut}, AV-SpeakerBench~\citep{avspeakerbench}, and VideoMME~\citep{videomme}. To assess the model's ability in real-world audio-visual interactive scenarios, we use Qualcomm IVD~\citep{qivd} as the benchmark for audio-query-based evaluation. Beyond understanding, we also evaluate the model's captioning capability on OmniCloze~\citep{omnicloze} and its tool-use ability on OmniGAIA~\citep{omnigaia}.

\subsubsection{Performance of Text$\to$Text}
We compare \method-Plus and \method-Flash with Qwen3.5-Plus-Instruct. As shown in Table \ref{tab:text_nonthink}, \method-Plus demonstrates text capabilities that are on par with its text-only counterpart across multiple dimensions, including knowledge, instruction following, long-context understanding, STEM, reasoning, and general agent tasks, highlighting its strong language ability. In particular, \method’s instruction-following performance is slightly better than the baseline. We believe that OPD and interaction-aligned RL have a positive effect on improving the instruction-following capabilities of an omni-model LLM.

\input{posttrain_tables/text.tex}

\subsubsection{Performance of Audio$\to$Text}
\input{posttrain_tables/omni35_audio2text.tex}

In Table \ref{tab:audio_benchmark}, we compare \method with Gemini-3.1 Pro in terms of audio-to-text performance. Compared to Gemini-3.1 Pro, \method exhibits superior performance on MMAU, MMSU, RUL-MuchoMusic, and SongFormBench, while achieving comparable results on MMAR, demonstrating its strong comprehension capabilities across multiple audio domains. Regarding end-to-end speech dialogue, \method significantly outperforms Gemini-3.1 Pro on VoiceBench and matches its performance on other benchmarks, further validating \method’s robust capabilities in end-to-end voice interaction. For S2TT and ASR, \method consistently outperforms Gemini-3.1 Pro, underscoring its superior translation and speech recognition performance across diverse languages, dialects, and domains.

\subsubsection{Performance of Vision $\to$ Text}
To comprehensively evaluate vision-to-text capabilities, we compare \flash and \plus with Qwen3.5-Plus-Instruct. As shown in Table \ref{tab:qwen3-omni-vl-results}, \plus achieves performance comparable to that of Qwen3.5-Plus-Instruct, while demonstrating stronger results on video understanding tasks involving both short and long videos. These findings highlight the strong dynamic visual perception ability of our model in real-world scenarios and suggest the effectiveness of joint video-audio training paradigms. Furthermore, we posit that audio-visual streams constitute the most naturalistic representation of real-world phenomena, wherein visual and auditory modalities are intrinsically coupled rather than independently processed.

\vspace{-.1in}

\input{posttrain_tables/vl}

\subsubsection{Performance of Audio-Visual Video$\to$Text}
We compare \method and Gemini-3.1 Pro across a diverse range of audio-visual tasks, as shown in Table~\ref{tab:qwen35-omni-omni-results}. For general understanding, \method achieves state-of-the-art performance on DailyOmni and obtains comparable results on AVUT. Our model also surpasses Gemini-3.1-Pro by a substantial margin on Qualcomm IVD, demonstrating its effectiveness in real-world audio-visual interactive scenarios.
Moreover, our model shows strong performance on captioning tasks. It can provide detailed audio, visual, and audio-visual captions. In this version, we also enhance the model's tool-use capability, achieving 57.2\% on OmniGAIA.
\input{posttrain_tables/va}

\subsection{Evaluation of X$\to$Speech}
In this section, we evaluate the speech generation capability of \method. Our evaluation mainly focuses on speech generation conditioned on text and prompt speech, following a zero-shot text-to-speech (TTS) setting. We study the model from four perspectives:

\begin{itemize}
    \item \textbf{Zero-Shot Speech Generation}: We evaluate content consistency, measured by WER, on SEED~\citep{seedtts}.
    \item \textbf{Multilingual Speech Generation}: We evaluate both content consistency and speaker similarity in zero-shot multilingual speech generation on the TTS multilingual test set~\citep{minimaxspeech} and an internal multilingual test set built on FLEURS~\citep{Conneau2022FLEURSFL}.
    \item \textbf{Cross-Lingual Speech Generation}: We evaluate content consistency in zero-shot cross-lingual speech generation on CV3-Eval~\citep{cosyvoice3}.
    \item \textbf{Custom-Voice Speech Generation}: We evaluate the stability of our speaker fine-tuned model on the TTS multilingual test set~\citep{minimaxspeech} and our internal multilingual test set.
\end{itemize}

\subsubsection{Evaluation of Zero-Shot Speech Generation}
We compare \method with state-of-the-art zero-shot TTS systems. As shown in Table~\ref{tab:zero_shot_speech_generation_table}, \method achieves highly competitive performance on the SEED-TTS benchmark, demonstrating strong content fidelity in zero-shot speech generation. These results reflect the effectiveness of our pretraining and continual pretraining pipeline in building robust speech generation and context modeling capabilities. Moreover, after RLHF optimization, \method further improves generation stability and naturalness, achieving the best performance on the \textit{test-en} split with a WER of 1.26.

\begin{table}[H]
\centering
\caption{\textbf{Zero-shot speech generation on the SEED-TTS test set.} Performance is measured by Word Error Rate (WER, $\downarrow$), where lower is better. The best results are highlighted in bold.}
\setlength{\tabcolsep}{2.6pt}
\begin{tabular}{@{}cll@{}}
\toprule
\textbf{Datasets} & \textbf{Model} & \textbf{Performance} \\
\midrule
\multicolumn{3}{c}{\textit{Content Consistency}} \\
\midrule 
\multirow{11}{*}{\begin{tabular}[c]{@{}c@{}}\textbf{SEED} \\ \textit{test-zh} | \textit{test-en} \end{tabular}}   
   & Seed-TTS\textsubscript{ICL}~\citep{seedtts} & 1.11 | 2.24  \\ 
   & Seed-TTS\textsubscript{RL}~\citep{seedtts} & 1.00 | 1.94  \\ 
   & MaskGCT~\citep{maskgct} & 2.27 | 2.62  \\ 
   & E2 TTS~\citep{e2tts} & 1.97 | 2.19  \\ 
   & F5-TTS~\citep{f5tts} & 1.56 | 1.83  \\ 
   & Spark TTS~\citep{sparktts} & 1.20 | 1.98  \\
   & CosyVoice 2~\citep{cosyvoice2} & 1.45 | 2.57  \\
   & CosyVoice 3~\citep{cosyvoice3} & \textbf{0.71} | 1.45  \\
   & MiniMax-Speech~\citep{minimaxspeech}            & 0.83 | 1.65  \\
   & MiMo-Audio-7B-Instruct~\citep{mimoaudio}            & 1.96 | 5.37  \\
   & Qwen2.5-Omni-7B~\citep{qwen2.5omni} & 1.42 | 2.33 \\
   & Qwen3-Omni-30B-A3B~\citep{qwen3omni} & 1.07 | 1.39 \\
   & \method-Plus & 0.99 | \textbf{1.26}  \\
\bottomrule
\end{tabular}
\label{tab:zero_shot_speech_generation_table}
\end{table}

\subsubsection{Evaluation of Multilingual Speech Generation}
\method supports speech generation in 29 languages. We compare its multilingual speech generation performance with two strong commercial systems, MiniMax-Speech and ElevenLabs. For the internal multilingual test set, we use GPT-4o-transcribe-2025-03-20 for automatic speech recognition.

As shown in Table~\ref{tab:multilingual_speech_generation_table} and Table~\ref{tab:in_multilingual_speech_generation_table}, \method achieves the lowest WER in 22 out of 29 evaluated languages on the multilingual test sets, outperforming the comparison systems by a clear margin in most cases. On the remaining languages, \method remains competitive with state-of-the-art systems. In addition to content consistency, \method also shows strong voice cloning fidelity. It obtains the highest speaker similarity scores in the majority of evaluated languages and consistently outperforms both MiniMax-Speech and ElevenLabs overall. These results suggest that \method effectively preserves speaker characteristics, such as timbre and prosodic style, while maintaining robust multilingual speech generation quality.

Furthermore, in Table~\ref{tab:in_multilingual_speech_generation_table}, we report results on our internal multilingual test set, covering an additional 9 languages. \method continues to achieve strong performance across all evaluated languages, indicating that its multilingual speech generation ability generalizes well beyond the public benchmark languages.

\begin{table}[t]
\centering
\footnotesize
\setlength{\tabcolsep}{4pt}
\caption{\textbf{Multilingual speech generation on the TTS multilingual test set.} Performance is measured by Word Error Rate (WER, $\downarrow$) for content consistency and cosine similarity (SIM, $\uparrow$) for speaker similarity. The best results are highlighted in bold.}
\begin{tabular}{@{}lcccccc@{}}
\toprule
\multirow{2}{*}{\textbf{Language}} & \multicolumn{3}{c}{\textbf{Content Consistency}}      & \multicolumn{3}{c}{\textbf{Speaker Similarity}}           \\ \cmidrule(l){2-7} 
                          & \tabincell{c}{\textbf{\method-Plus}}       & \textbf{MiniMax} & \textbf{ElevenLabs} & \tabincell{c}{\textbf{\method-Plus}}       & \textbf{MiniMax} & \textbf{ElevenLabs} \\ \midrule
Chinese    & \textbf{0.695} & 2.252          & 16.026         & \textbf{0.800}   & 0.780           & 0.677      \\
English    & \textbf{0.631} & 2.164          & 0.756          & \textbf{0.833} & 0.756          & 0.613      \\
German     & \textbf{0.447} & 1.906          & 0.572          & \textbf{0.757} & 0.733          & 0.614      \\
Italian    & \textbf{0.503} & 1.543          & 1.743          & \textbf{0.785} & 0.699          & 0.679      \\
Portuguese & \textbf{1.221} & 1.877          & 1.331          & 0.792          & \textbf{0.805} & 0.711      \\
Spanish    & \textbf{0.862} & 1.029          & 1.084          & \textbf{0.797} & 0.762          & 0.615      \\
Japanese   & \textbf{3.479} & 3.519          & 10.046         & \textbf{0.788} & 0.776          & 0.738      \\
Korean     & \textbf{1.458} & 1.747          & 1.865          & 0.747          & \textbf{0.776} & 0.700      \\
French     & \textbf{2.430} & 4.099          & 5.216          & \textbf{0.730} & 0.628          & 0.535      \\
Russian    & \textbf{3.182} & 4.281          & 3.878          & \textbf{0.790} & 0.761          & 0.676      \\
Thai       & \textbf{2.170} & 2.701          & 73.936         & 0.788          & \textbf{0.800} & 0.588      \\
Indonesian & \textbf{0.823} & 1.237          & 1.059          & \textbf{0.780} & 0.729          & 0.660      \\
Arabic     & 2.602          & \textbf{1.665} & 1.666          & \textbf{0.745} & 0.736          & 0.706      \\
Vietnamese & 1.143          & \textbf{0.880} & 73.415         & \textbf{0.767} & 0.743          & 0.369      \\
Turkish    & 0.938          & 1.520          & \textbf{0.699} & 0.747          & \textbf{0.779} & 0.596      \\
Finnish    & \textbf{2.784} & 4.666          & 2.964          & \textbf{0.859} & 0.835          & 0.759      \\
Polish     & 1.427          & 1.415          & \textbf{0.766} & \textbf{0.839} & 0.802          & 0.729      \\
Hindi      & 6.444          & 6.962          & \textbf{5.827} & 0.797          & \textbf{0.818} & 0.730      \\
Dutch      & 1.238          & 1.143          & \textbf{0.803} & \textbf{0.762} & 0.738          & 0.680      \\
Czech      & 2.929          & 3.875          & \textbf{2.108} & \textbf{0.802} & 0.796          & 0.685      \\ 
\bottomrule
\end{tabular}
\label{tab:multilingual_speech_generation_table}
\end{table}

\begin{table}[H]
\centering
\footnotesize
\setlength{\tabcolsep}{4pt}
\caption{\textbf{Multilingual speech generation on the internal multilingual test set.} Performance is measured by Word Error Rate (WER, $\downarrow$) for content consistency and cosine similarity (SIM, $\uparrow$) for speaker similarity. The best results are highlighted in bold.}
\begin{tabular}{@{}lcccc@{}}
\toprule
\multirow{2}{*}{\textbf{Language}} & \multicolumn{2}{c}{\textbf{Content Consistency}}      & \multicolumn{2}{c}{\textbf{Speaker Similarity}}           \\ \cmidrule(l){2-5} 
                          & \tabincell{c}{\textbf{\method-Plus}}       & \textbf{Ground Truth} & \tabincell{c}{\textbf{\method-Plus}}       & \textbf{Ground Truth} \\ \midrule
Urdu      & \textbf{14.819} & 17.822       & 0.775 & -            \\
Tagalog   & \textbf{5.193}  & 6.885        & 0.870 & -            \\
Swedish   & \textbf{3.760}  & 4.813        & 0.822 & -            \\
Danish    & \textbf{3.636}  & 6.403        & 0.775 & -            \\
Hebrew    & \textbf{7.860}  & 16.178       & 0.760 & -            \\
Icelandic & \textbf{10.244} & 11.451       & 0.764 & -            \\
Malay     & \textbf{3.142}  & 4.628        & 0.794 & -            \\
Norwegian & \textbf{3.613}  & 4.442        & 0.825 & -            \\
Persian   & \textbf{11.113} & 14.469       & 0.800 & -            \\ 
\bottomrule
\end{tabular}
\label{tab:in_multilingual_speech_generation_table}
\end{table}

\subsubsection{Evaluation of Cross-Lingual Speech Generation}
Beyond multilingual voice cloning, \method also supports cross-lingual voice cloning, where the model is required to preserve speaker identity while generating speech in a different target language. We evaluate this capability on the Cross-Lingual benchmark and compare against the CosyVoice series as well as Qwen3-Omni-30B-A3B.

In Table~\ref{tab:cross_lingual_speech_generation_table}, we report the mixed error rate (WER for English and CER for the other languages) across different source--target language pairs. Overall, \method achieves the best performance in 10 out of 12 evaluated directions and sets a new state of the art on most English-, Japanese-, and Korean-targeted pairs. In particular, for \textit{zh-to-ko}, \method reduces the error rate from 14.4 to 4.03 compared with CosyVoice3, corresponding to an approximately 72\% relative reduction. \method also performs strongly on commonly used language pairs such as \textit{zh-to-en} and \textit{en-to-zh}, indicating better content consistency under cross-lingual generation. These results demonstrate that \method generalizes effectively across language boundaries while preserving target linguistic accuracy.

\begin{table}[t]
\centering
\caption{\textbf{Cross-lingual speech generation on the Cross-Lingual benchmark.} Performance is measured by mixed error rate (WER for English and CER for the other languages, $\downarrow$). The best results are highlighted in bold.}
\resizebox{0.95\textwidth}{!}{%
\begin{tabular}{@{}lcccc@{}}
\toprule
\textbf{Language} & \multicolumn{1}{c}{\textbf{\method-Plus}} & \multicolumn{1}{c}{\textbf{Qwen3-Omni-30B-A3B}} & \multicolumn{1}{c}{\textbf{CosyVoice3}} & \multicolumn{1}{c}{\textbf{CosyVoice2}} \\ \midrule
English-to-Chinese & \textbf{4.86} & 5.37 & 5.09 & 13.5 \\
Japanese-to-Chinese & 3.55 & 3.32 & \textbf{3.05} & 48.1 \\
Korean-to-Chinese & \textbf{0.84} & 0.99 & 1.06 & 7.70 \\
Chinese-to-English & \textbf{2.18} & 2.76 & 2.98 & 6.47 \\
Japanese-to-English & \textbf{2.18} & 3.31 & 4.20 & 17.1 \\
Korean-to-English & \textbf{2.51} & 3.34 & 4.19 & 11.2 \\
Chinese-to-Japanese & \textbf{5.92} & 8.29 & 7.08 & 13.1 \\
English-to-Japanese & \textbf{5.12} & 7.53 & 6.80 & 14.9 \\
Korean-to-Japanese & \textbf{2.16} & 4.24 & 3.93 & 5.86 \\
Chinese-to-Korean & \textbf{4.03} & 5.13 & 14.4 & 24.8 \\
English-to-Korean & \textbf{3.72} & 4.96 & 5.87 & 21.9 \\
Japanese-to-Korean & \textbf{5.12} & 6.23 & 7.92 & 21.5 \\ 
\bottomrule
\end{tabular}
}
\label{tab:cross_lingual_speech_generation_table}
\end{table}
\vspace{-0.1in}
\subsubsection{Evaluation of Custom-Voice Speech Generation}

We evaluate the custom-voice speech generation capability of \method in multilingual settings. We compare \method with several strong commercial systems accessed through their official APIs in March 2026, including ElevenLabs Multilingual v2 (9YHcvj6GT2YYXdXww), Gemini-2.5 Pro-Preview-TTS (Achernar), GPT-Audio-2025-08-28 (Alloy), and MiniMax-Speech-2.8-HD (English\_expressive\_narrator).

\begin{table}[H]
\centering
\caption{\textbf{Custom-voice multilingual speech generation on the multilingual test set.} Performance is measured by Word Error Rate (WER, $\downarrow$). The best results are highlighted in bold.}
\begin{tabular}{@{}lccccc@{}}
\toprule
\textbf{Language}     & \textbf{\method-Plus} & \textbf{ElevenLabs} & \textbf{Gemini-2.5 Pro} & \textbf{GPT-Audio}   & \textbf{MiniMax} \\ \midrule
Chinese    & \textbf{0.785}     & 3.801              & 1.890          & 0.829           & 0.786            \\
English    & \textbf{0.839}     & 1.126              & 0.953          & 1.050           & 1.429            \\
German     & \textbf{0.182}     & 0.500              & 0.509          & 0.558           & 1.581            \\
Italian    & \textbf{0.458}     & 0.513              & 0.991          & 0.769           & 1.063            \\
Portuguese & 1.581              & \textbf{1.109}     & 2.050          & 1.506           & 1.240            \\
Spanish    & 0.768              & \textbf{0.520}     & 0.891          & 0.936           & 0.691            \\
Japanese   & \textbf{3.306}     & 11.685             & 4.420          & 4.317           & 4.254            \\
Korean     & \textbf{1.309}     & 3.981              & 4.110          & 3.999           & 3.635            \\
French     & 2.724              & \textbf{2.574}     & 3.284          & 2.809           & 3.439            \\
Russian    & 4.723              & 4.324              & 3.858          & 4.346           & 3.529            \\
Thai       & \textbf{1.653}     & 114.813            & 2.539          & 4.430           & 1.811            \\
Indonesian & 1.596              & 6.094              & \textbf{1.498} & 2.362           & 1.585            \\
Arabic     & \textbf{3.183}     & 5.400              & 5.525          & 5.326           & 3.309            \\
Vietnamese & 1.320              & 82.849             & 1.699          & 1.854           & \textbf{1.058}   \\
Turkish    & 1.309              & \textbf{0.551}     & 2.237          & 1.389           & 0.652            \\
Finnish    & 4.039              & \textbf{2.522}     & 5.331          & 3.270           & 2.939            \\
Polish     & 1.462              & \textbf{0.733}     & 1.622          & 1.737           & 0.833            \\
Hindi      & 6.776              & 6.388              & 6.596          & 7.191           & \textbf{6.146}   \\
Dutch      & 1.135              & 1.005              & \textbf{0.973} & 1.561           & 1.406            \\
Czech      & 3.769              & 1.916              & 3.380          & 2.859           & \textbf{1.766}   \\
Urdu       & 14.916             & 12.970             & 14.141         & \textbf{13.362} & 24.151           \\
Tagalog    & \textbf{5.090}     & 5.473              & 6.784          & 5.352           & 5.674            \\
Swedish    & 3.588              & 3.132              & 3.196          & 2.898           & \textbf{2.833}   \\
Danish     & 7.183              & \textbf{2.604}     & 3.876          & 3.846           & 4.951            \\
Hebrew     & 7.680              & 102.018            & \textbf{4.459} & 5.328           & 8.161            \\
Icelandic  & 10.322             & 25.110             & \textbf{6.348} & 9.648           & 33.431           \\
Malay      & 3.738              & 6.448              & 3.731          & \textbf{3.406}  & 3.955            \\
Norwegian  & 5.576              & 7.351              & 4.304          & \textbf{3.400}  & 9.492            \\
Persian    & \textbf{12.140}    & 20.564             & 12.620         & 13.202          & 12.722           \\ 
\bottomrule
\end{tabular}
\label{tab:single_speech_generation_table}
\end{table}

As shown in Table~\ref{tab:single_speech_generation_table}, although \method is fine-tuned only on monolingual data, it demonstrates strong cross-lingual generalization in custom-voice speech generation. The model is able to transfer the target speaker characteristics to all 29 evaluated languages while maintaining stable generation quality. Overall, \method achieves the best WER in 10 languages and remains competitive in many others. In particular, it shows clear advantages in several challenging languages, including Japanese (3.306) and Korean (1.309), indicating strong intelligibility under cross-lingual voice transfer. These results suggest that \method can generate custom-voice speech with robust linguistic fidelity across a wide range of languages.


%% file: posttrain_tables/text.tex
\begin{table}[H]
\centering
\caption{\textbf{Text $\to$ Text performance of \method and Qwen3.5-Plus-Instruct. The highest
scores are shown in bold.}}
\vspace{-.1in}
\label{tab:text_nonthink}
\begin{tabular}{@{}lccc@{}}
\toprule
\multicolumn{1}{c}{\textbf{Datasets}} & \textbf{Qwen3.5-Plus-Instruct} & \textbf{\method-Flash} & \textbf{\method-Plus} \\ \midrule
\multicolumn{4}{c}{\textit{Knowledge}}                                                                                              \\ \midrule
MMLU-Pro                              & \textbf{86.8}                    & 79.9                        & 85.9                       \\
MMLU-Redux                            & \textbf{94.3}                    & 90.0                        & 94.2                       \\
SuperGPQA                             & \textbf{67.4}                    & 54.9                        & 66.4                       \\
C-Eval                                & \textbf{92.3}                    & 86.0                        & 92.0                       \\ \midrule
\multicolumn{4}{c}{\textit{Instruction Following}}                                                                                  \\ \midrule
IFEval                                & \textbf{89.7}                    & 85.2                        & \textbf{89.7}              \\
IFBench                               & 51.1                             & 38.4                        & \textbf{52.6}              \\ \midrule
\multicolumn{4}{c}{\textit{Long Context}}                                                                                           \\ \midrule
AA-LCR                                & \textbf{62.0}                    & 46.0                        & 57.0                       \\
LongBench v2                          & \textbf{60.2}                    & 46.4                        & 59.6                       \\ \midrule
\multicolumn{4}{c}{\textit{STEM}}                                                                                                   \\ \midrule
GPQA                                  & \textbf{85.9}                    & 76.4                        & 83.9                       \\ \midrule
\multicolumn{4}{c}{\textit{Reasoning}}                                                                                              \\ \midrule
LiveCodeBench v6                      & \textbf{67.1}                    & 56.6                        & 65.6                       \\
HMMT Nov 25                           & \textbf{86.2}                    & 59.0                        & 84.4                       \\
IMOAnswerBench                        & \textbf{68.3}                    & 51.5                        & 65.5                       \\ \midrule
\multicolumn{4}{c}{\textit{General Agent}}                                                                                          \\ \midrule
BFCL-V4                               & \textbf{66.1}                    & 55.3                        & 63.3                       \\
TAU2Bench                             & \textbf{82.7}                    & 78.0                        & 81.0                       \\ \bottomrule
\end{tabular}
\end{table}

%% file: posttrain_tables/omni35_audio2text.tex
\begin{table*}[ht]
\centering
\caption{\textbf{Audio benchmark comparison across Gemini-3.1 Pro, \method-Flash, and \method-Plus. For most benchmarks, higher is better. For ASR benchmarks, lower WER is better. Best results are shown in bold.}}
\small
\label{tab:audio_benchmark}
\begin{threeparttable}
\resizebox{\textwidth}{!}{%
\begin{tabular}{lccc}
\toprule
\textbf{Datasets} & \textbf{Gemini-3.1 Pro} & \textbf{\method-Flash} & \textbf{\method-Plus} \\
\midrule

\multicolumn{4}{c}{\textit{Audio Understanding ($\uparrow$)}} \\
\midrule
MMAU & 81.1 & 80.4 & \textbf{82.2} \\
MMAR & \textbf{83.7} & 74.0 & 80.0 \\
MMSU & 81.3 & 72.2 & \textbf{82.8} \\
RUL-MuchoMusic & 59.6 & 60.5 & \textbf{72.4} \\
SongFormBench-HarmonixSet$_{\text{(acc|hr.5f|hr3f)}}$\tnote{a} & 75.6 | 46.8 | 77.9 & 80.6 | 67.8 | 83.4 & \textbf{81.1 | 72.9 | 85.3} \\
SongFormBench-CN$_{\text{(acc|hr.5f|hr3f)}}$\tnote{a} & 78.1 | 43.2 | 71.9 & 86.7 | \textbf{66.4} | \textbf{84.6} & \textbf{87.1} | 65.7 | 84.2 \\

\midrule
\multicolumn{4}{c}{\textit{Dialogue} ($\uparrow$)} \\
\midrule
VoiceBench & 88.9 & 87.8 & \textbf{93.1} \\
URO-Bench-pro$_{\text{(U|R|O)}}$\tnote{b} & \textbf{69.1} | 84.0 | 99.2 & 64.1 | 83.8 | 98.7 & 66.3 | \textbf{86.3} | \textbf{99.8} \\
SpeechRole & \textbf{124.2} & 119.8 & 123.5 \\
WildSpeech-Bench & \textbf{76.3} & 72.2 & 75.4 \\

\midrule
\multicolumn{4}{c}{\textit{S2TT} ($\uparrow$)} \\
\midrule
Fleurs$_{\text{xx}\leftrightarrow\text{zh (top59)}}$\tnote{c} & 29.5 & 26.9 & \textbf{30.2} \\
Fleurs$_{\text{xx}\leftrightarrow\text{en (top59)}}$\tnote{c} & 34.6 & 32.0 & \textbf{35.4} \\
Fleurs$_{\text{xx}\leftrightarrow\text{zh/en (top59)}}$\tnote{c} & 32.1 & 29.4 & \textbf{32.8} \\

\midrule
\multicolumn{4}{c}{\textit{ASR (WER$\downarrow$)}} \\
\midrule
Fleurs$_{\text{(top60)}}$ & 7.32 & 10.75 & \textbf{6.55} \\
CV15$_{\text{(zh|yue|zh-tw)}}$ & 8.59 | 13.40 | 6.78 & 4.25 | 3.45 | 2.68 & \textbf{3.46 | 1.95 | 2.27} \\
CV15$_{\text{(en)}}$ & 8.73 & 5.90 & \textbf{4.83} \\
Librispeech$_{\text{(clean|other)}}$ & 3.36 | 4.41 & 1.30 | 2.43 & \textbf{1.11 | 2.23} \\
Weneetspeech$_{\text{(net|meeting)}}$ & 11.53 | 14.21 & 4.41 | \textbf{5.51} & \textbf{4.30} | 5.84 \\
Kespeech & 23.67 & 4.47 & \textbf{3.46} \\
MIR-1K$_{\text{(vocal-only)}}$\tnote{d} & 8.76 & 4.94 & \textbf{4.56} \\
Opencpop & 6.83 & \textbf{1.11} & 1.49 \\

\bottomrule
\end{tabular}
}
\begin{tablenotes}
    \footnotesize
    \item[a] SongFormBench: We use a unified prompt defining an SRT-like output timestamp format and a closed vocabulary for evaluation. The vocabulary follows the \textit{SongForm-HX-8Class} specified in the official codebase.
    \item[b] URO-Bench-Pro: We use the pro track of URO-Bench and denote the three evaluation dimensions as follows: U for Understanding, R for Reasoning, and O for Oral Conversation. We use GenStyle-en, GenStyle-zh, Multilingual tasks for oral dimension.
    \item[c] Fleurs: The top59 languages are English, Chinese, Cantonese, Korean, Japanese, Vietnamese, Thai, Malay, German, Russian, Italian, French, Spanish, Portuguese, Dutch, Indonesian, Turkish, Arabic, Polish, Hindi, Urdu, Filipino, Persian, Czech, Greek, Swedish, Hebrew, Danish, Finnish, Norwegian, Icelandic, Bengali, Punjabi, Javanese, Marathi, Swahili, Ukrainian, Gujarati, Kannada, Azerbaijani, Malayalam, Cebuano, Romanian, Hungarian, Bulgarian, Belarusian, Catalan, Tamil, Croatian, Bosnian, Slovak, Galician, Kyrgyz, Macedonian, Slovenian, Latvian, Estonian, and Asturian; compared with the top60 list, Afrikaans is excluded because the Fleurs S2TT test set does not cover this language.
    \item[d] MIR-1K: Transcription is converted into Simplified Chinese.
\end{tablenotes}
\end{threeparttable}
\end{table*}

%% file: posttrain_tables/vl.tex
\begin{table}[htb]
\centering
\caption{\textbf{Vision $\to$ Text performance of \method and Qwen3.5-Plus-Instruct. The highest scores are shown in bold.}}
\vspace{-.1in}
\label{tab:qwen3-omni-vl-results}
\begin{tabular}{@{}lccc@{}}
\toprule
\textbf{Datasets}  & \tabincell{c}{\textbf{Qwen3.5-Plus-Instruct}} & \tabincell{c}{\textbf{\flash}} & \tabincell{c}{\textbf{\plus}} \\
\midrule
\multicolumn{4}{c}{\textit{STEM and Puzzle}} \\
\midrule
MMMU & 81.0 & 76.9 & 80.1 \\
MMMU-Pro & 73.8 & 68.2 & 73.9 \\
MathVision & 73.6 & 65.4 & 73.0 \\
Mathvista (mini) & 86.9 & 82.9 & 86.1 \\
DynaMath & 84.2 & 79.3 & 83.8 \\
ZEROBench & 6 & 1 & 5 \\
ZEROBench\_sub & 31.1 & 26.0 & 34.4 \\
\midrule
\multicolumn{4}{c}{\textit{General VQA}} \\
\midrule
RealWorldQA & 79.1 & 77.5 & 84.1 \\
MMStar & 80.3 & 75.7 & 79.4 \\
MMBench\textsubscript{EN-DEV-v1.1} & 93.8 & 88.8 & 92.8 \\
SimpleVQA & 66.1 & 54.4 & 65.3 \\
\midrule
\multicolumn{4}{c}{\textit{Text Recognition and Document Understanding}} \\
\midrule
CharXiv (RQ) & 74.2 & 64.4 & 72.5 \\
CC-OCR & 83.0 & 80.8 & 83.4 \\
AI2D\_TEST & 92.1 & 89.0 & 91.2 \\
MMLongBench-Doc & 59.7 & 53.6 & 57.5 \\
OCRBench & 91.4 & 89.1 & 91.3 \\
\midrule
\multicolumn{4}{c}{\textit{Spatial Intelligence}} \\
\midrule
ERQA & 53.8 & 50.0 & 54.8 \\
CountBench & 95.1 & 88.2 & 95.1 \\
RefCOCO(avg) & 95.2 & 92.6 & 95.0 \\
ODInW13 & 50.3 & 46.8 & 49.5 \\
EmbSpatialBench & 83.4 & 82.7 & 85.4 \\
\midrule
\multicolumn{4}{c}{\textit{Video Understanding}} \\
\midrule
VideoMME\textsubscript{(w/o sub.)} & 81.0 & 77.0 & 81.9 \\
MLVU\textsubscript{(M-Avg)} & 85.1 & 81.9 & 86.8 \\
MVBench & 76.7 & 70.8 & 79.0 \\
LVBench & 68.6 & 65.7 & 71.2 \\
MMVU & 67.1 & 62.7 & 67.5 \\
MME-VideoOCR & 74.2 & 70.5 & 77.0 \\
\midrule
\multicolumn{4}{c}{\textit{Medical VQA}} \\
\midrule
SLAKE & 82.8 & 73.1 & 84.7 \\
PMC-VQA & 62.4 & 58.7 & 62.7 \\
MedXpertQA-MM & 55.3 & 44.8 & 54.7 \\
\bottomrule
\end{tabular}
\end{table}


%% file: posttrain_tables/va.tex
\begin{table}[htbp]
\centering
\caption{\textbf{Audio-Visual $\to$ Text performance of \method and Gemini-3.1-Pro. The highest scores are shown in bold.}}
\vspace{-1mm}
\label{tab:qwen35-omni-omni-results}
\adjustbox{center=\textwidth}{
\small
\setlength{\tabcolsep}{18pt}
\begin{threeparttable}
\begin{tabular}{@{}lccc@{}}
\toprule
\textbf{Datasets} & \tabincell{c}{\textbf{Gemini-3.1 Pro}} & \tabincell{c}{\textbf{\method-Flash}} & \tabincell{c}{\textbf{\method-Plus}} \\
\midrule
\multicolumn{4}{c}{\textit{Text Query QA}} \\
\midrule
DailyOmni                            & 82.7          & 81.8          & \textbf{84.6} \\
WorldSense                           & \textbf{65.5} & 57.9          & 62.8          \\
AVUT                                 & \textbf{85.6} & 81.4          & 85.0          \\
AV-SpeakerBench                      & \textbf{75.1} & 65.2          & 71.3          \\
VideoMME\textsubscript{w/ audio}\tnote{a} & \textbf{89.0} & 79.3     & 83.7          \\
\midrule
\multicolumn{4}{c}{\textit{Audio Query QA}} \\
\midrule
Qualcomm IVD                  & 66.2          & 66.3          & \textbf{68.5} \\
\midrule
\multicolumn{4}{c}{\textit{Caption}} \\
\midrule
Omni-Cloze                           & 57.2          & 63.0          & \textbf{64.8} \\
\midrule
\multicolumn{4}{c}{\textit{Agent (Tool Use)}} \\
\midrule
OmniGAIA\tnote{b}                    & \textbf{68.9} & 33.9          & 57.2          \\
\bottomrule
\end{tabular}
\begin{tablenotes}
    \footnotesize
    \item[a] VideoMME is evaluated with \texttt{use\_audio\_in\_video=True}.
    \item[b] OmniGAIA is evaluated without a thinking prompt and without \texttt{<answer>} formatting. All results are evaluated using DeepSeek-V3.2-Thinking as the judge.
\end{tablenotes}
\end{threeparttable}
}
\end{table}

%% file: content/conclusion.tex
\section{Conclusion}
\label{sec:conclusion}
In this work, we present \method, a fully omnimodal large language model that unifies understanding, reasoning, generation, and action across text, images, audio, and audio-visual inputs. Built on the Thinker--Talker framework, \method introduces efficient Hybrid-Attention MoE architectures, 256k long-context modeling, improved streaming speech generation with multi-codebook codec prediction and ARIA, and substantially expanded multilingual speech support. These advances enable three key capabilities: controllable audio-visual captioning, comprehensive real-time interaction, and native omnimodal agentic behavior through autonomous tool use and audio-visual code generation. Empirically, \method achieves state-of-the-art or highly competitive performance across a broad range of audio and audio-visual benchmarks, while maintaining the strong text and vision capabilities of same-scale Qwen models. These results suggest that scaling native omnimodal training can produce unified systems that not only perceive and reason across modalities, but also interact and act in real time. We hope \method provides a strong foundation for future research on general-purpose omnimodal agents.

%% file: content/authors.tex
\section{Authors}

\textbf{Core Contributors\footnotemark}
\begin{multicols}{6}
{\fontsize{9}{10}\selectfont
\input{content/contributor/core_contributor}
}
\end{multicols}
\footnotetext{Alphabetical order. * denotes the corresponding author.}

\textbf{Contributors\footnotemark[\value{footnote}]}
\begin{multicols}{6}
{\fontsize{9}{10}\selectfont
\input{content/contributor/contributor}
}
\end{multicols}

%% file: content/contributor/core_contributor.tex
Bing Han\\
Baosong Yang\\
Bin Zhang\\
Bo Zheng\\
Dayiheng Liu\\
Fan Zhou\\
Hongkun Hao\\
Hangrui Hu\\
Jin Xu$^*$\\
Jianxin Yang\\
Jingren Zhou\\
Keqin Chen\\
Le Yu\\
Mingkun Yang\\
Peng Wang\\
Pei Zhang\\
Qize Yang\\
Rui Men\\
Ruiyang Xu\\
Shuai Bai\\
Sibo Song\\
Ting He\\
Xize Cheng\\
Xuejing Liu\\
Xingzhang Ren\\
Xian Shi\\
Xiong Wang\\
Xinyu Zhang\\
Xinfa Zhu\\
Yunfei Chu\\
Yuanjun Lv\\
Yuchong Sun\\
Yongqi Wang\\
Yuxuan Wang\\
Yang Zhang\\
Zhifang Guo\\
Zishan Guo\\
Ziyang Ma\\

%% file: content/contributor/contributor.tex
Andong Chen\\
Anfeng Li\\
An Yang\\
Bei Chen\\
Bin Lin\\
Bingshen Mu\\
Bohan Wang\\
Buxiao Wu\\
Bowen Xu\\
Beichen Zhang\\
Cheng Chen\\
Chang Gao\\
Chengen Huang\\
Chenyang Le\\
Chenhao Li\\
Chenglong Liu\\
Chenxu Lv\\
Chen Qiang\\
Chenfei Wu\\
Chenhan Yuan\\
Chengruidong Zhang\\
Chujie Zheng\\
Daren Chen\\
Dake Guo\\
Fei Huang\\
Gaoji Liu\\
Guangdong Zhou\\
Hao Ge\\
Huiqiang Jiang\\
Haoran Lian\\
Hongjian Tu\\
Hao Yu\\
Hang Zhang\\
Hao Zhou\\
Haiquan Zhao\\
Humen Zhong\\
Jiawei Chen\\
Jian Guan\\
Jiayi Leng\\
Jiahao Li\\
Junrong Lin\\
Jiawei Liu\\
Jialong Tang\\
Jun Tang\\
Jianhong Tu\\
Jianqiang Wan\\
Jinxi Wei\\
Jianwei Zhang\\
Jing Zhou\\
Kai Dang\\
Kangxiang Xia\\
Kun Yan\\
Kexin Yang\\
Lianghao Deng\\
Lulu Hu\\
Linhan Ma\\
Lingchen Meng\\
Lei Xie\\
Laiwen Zheng\\
Miao Hong\\
Mei Li\\
Mingcheng Li\\
Mingze Li\\
Minsheng Li\\
Minghao Wu\\
Mingfeng Xue\\
Na Ni\\
Peng Liu\\
Peng Wang\\
Pengfei Wang\\
Peiyang Zhang\\
Qidong Huang\\
Qingfeng Lan\\
Qintong Li\\
Que Shen\\
Qiuyue Wang\\
Qin Zhu\\
Ruisheng Cao\\
Rongyao Fang\\
Rui Hu\\
Ruibin Yuan\\
Song Chen\\
Su Hao\\
Shen Li\\
Shixuan Liu\\
Shurui Li\\
Siqi Zhang\\
Tianyi Tang\\
Tingyu Xia\\
Wei Ding\\
Wenbin Ge\\
Weizhou Shen\\
Wei Wang\\
Wentao Yao\\
Xi Chen\\
Xiaotong Chen\\
Xionghui Chen\\
Xiaodong Deng\\
Xudong Guo\\
Xin Le\\
Xiao Li\\
Xie Chen\\
Xinyao Niu\\
Xuancheng Ren\\
Xuechun Wang\\
Xuwu Wang\\
Xingzhe Wu\\
Xipin Wei\\
Xiao Xu\\
Xian Yang\\
Yuxuan Cai\\
Yizhong Cao\\
Yilei Chen\\
Yuxiang Chen\\
Yiming Dong\\
Yang Fan\\
Yanpeng Li\\
Yucheng Li\\
Yang Liu\\
Yantao Liu\\
Yuqiong Liu\\
Yuxuan Liu\\
Yuyan Luo\\
Yubo Ma\\
Yang Su\\
Yuezhang Wang\\
Yuhao Wang\\
Yi Wu\\
Yunbao Wu\\
Yu Xi\\
Yi Zhang\\
Yichang Zhang\\
Yinger Zhang\\
Yuxiang Zheng\\
Zeyu Cui\\
Ziwei Ji\\
Ziyue Jiang\\
Zhaohai Li\\
Zheng Li\\
Zhi Li\\
Zihan Qiu\\
Zekun Wang\\
Zhihai Wang\\
Zhenghao Xing\\
Zhibo Yang\\
Zhuorui Ye\\
Zhenru Zhang\\
Zhipeng Zhou\\
Zhengyang Zhuge\\

%% file: content/appendix.tex
\section{Appendix}
\label{sec:appendix}

\subsection{Detailed Multilingual Evaluation Results}
\label{sec:mutilingual-results}

\paragraph{Multilingual ASR.}
As presented in Table~\ref{tab:multilingual_asr_detail_table}, Qwen3.5-Omni demonstrates superior speech recognition capabilities compared to state-of-the-art competitors on the FLEURS test set. Qwen3.5-Omni-Plus achieves the lowest average WER of 6.6\%, outperforming both Gemini-3.1-Pro (7.3\%) and GPT-4o-Transcribe (10.4\%). It secures the best performance in the majority of languages, with particularly significant margins in complex tonal and low-resource languages such as Cantonese (2.2\% vs. 6.3\% for Gemini-3.1-Pro), Thai, and Vietnamese. Meanwhile, Qwen3.5-Omni-Flash offers a highly efficient alternative, achieving an average WER of 10.8\% that remains competitive against Gemini-3-Flash (10.5\%). Notably, Qwen3.5-Omni-Flash exhibits exceptional robustness in challenging scenarios, drastically reducing errors in Cantonese (3.1\% vs. 10.8\% for Gemini-3-Flash) and maintaining strong performance in Japanese and Korean, thereby highlighting its advantage for high-value Asian language pairs.

\paragraph{Multilingual Translation.}
As shown in Tables~\ref{tab:multilingual_speech_translation_2xx_table} and~\ref{tab:multilingual_speech_translation_xx2_table},  the Qwen3.5-Omni series demonstrates distinct advantages over state-of-the-art competitors on the FLEURS test set, particularly in Asian languages and specific high-resource pairs. Qwen3.5-Omni-Plus exhibits comprehensive superiority over Gemini-3.1-Pro in the many-to-many directions (en2xx/zh2xx), achieving higher average BLEU scores in both English-to-XX (33.8 vs. 31.8) and Chinese-to-XX (21.4 vs. 19.6). It also leads in key xx2en pairs such as Portuguese (49.4 vs. 47.7) and Indonesian (45.7 vs. 45.1). Although Gemini-3.1-Pro holds a slight edge in overall xx2zh averages, Qwen3.5-Omni-Plus significantly outperforms it in critical Asian languages, including Cantonese (+15.6 BLEU), Korean, and Japanese. Similarly, Qwen3.5-Omni-Flash shows targeted strengths against Gemini-3-Flash. While maintaining competitive general performance, it vastly surpasses Gemini in Cantonese translation across all directions (e.g., 37.5 vs. 22.4 in xx2zh and 37.3 vs. 26.7 in en2xx) and delivers better results in Japanese and Korean xx2zh tasks. These results underscore Qwen3.5-Omni’s robust optimization for complex Asian linguistic structures and key regional languages.

\begin{table}[ht]
\centering
\small
\caption{Multilingual ASR performance on the FLEURS test set. Results are reported using Word Error Rate (WER, ↓), where lower values indicate better performance. For italicized languages, Character Error Rate (CER, ↓) is reported instead. Compared with competing models, Qwen3.5-Omni-Plus achieves the best results on the majority of languages. The best results are highlighted in bold.}
\setlength\tabcolsep{8pt}
\begin{tabular}{lccccc}
\toprule
\textbf{Language} 
& \begin{tabular}[c]{@{}c@{}}\textbf{Qwen3.5-}\\ \textbf{Omni-Plus}\end{tabular} 
& \begin{tabular}[c]{@{}c@{}}\textbf{Qwen3.5-}\\ \textbf{Omni-Flash}\end{tabular} 
& \begin{tabular}[c]{@{}c@{}}\textbf{Gemin-3.1-}\\ \textbf{Pro}\end{tabular} 
& \begin{tabular}[c]{@{}c@{}}\textbf{GPT-4o-}\\ \textbf{Transcribe}\end{tabular} 
& \begin{tabular}[c]{@{}c@{}}\textbf{Gemini-3-}\\ \textbf{Flash}\end{tabular} \\
\midrule
\textit{Chinese} & 2.9 & 2.9 & 3.6 & \textbf{2.6} & 4.6 \\
English & 3.2 & 3.7 & \textbf{2.7} & 3.2 & 2.9 \\
\textit{Cantonese} & \textbf{2.2} & 3.1 & 6.3 & 5.2 & 10.8 \\
Arabic & 11.7 & 13.6 & \textbf{9.2} & 13.0 & 10.1 \\
German & \textbf{2.0} & 2.5 & 2.7 & 2.3 & 3.3 \\
French & \textbf{2.6} & 3.3 & 3.7 & 3.7 & 3.9 \\
Spanish & \textbf{2.2} & 2.4 & 2.5 & 2.3 & 2.7 \\
Portuguese & \textbf{2.1} & 2.2 & 2.6 & 2.3 & 2.9 \\
Indonesian & \textbf{1.6} & 2.4 & 2.5 & 3.5 & 2.8 \\
Italian & \textbf{0.8} & 1.0 & 1.1 & 1.4 & 1.8 \\
\textit{Korean} & \textbf{1.7} & 2.1 & 2.0 & 2.1 & 2.4 \\
Russian & \textbf{3.1} & 3.6 & 3.4 & 3.7 & 3.9 \\
\textit{Thai} & \textbf{2.8} & 3.2 & 4.3 & 4.9 & 4.5 \\
\textit{Vietnamese} & \textbf{1.9} & 2.5 & 2.5 & 3.5 & 3.5 \\
\textit{Japanese} & \textbf{1.9} & 2.5 & 2.3 & 3.0 & 3.4 \\
Turkish & \textbf{3.1} & 4.4 & 3.8 & 4.2 & 4.4 \\
Hindi & 9.7 & 9.9 & \textbf{4.5} & 12.0 & 5.6 \\
Malay & \textbf{2.7} & 4.2 & 3.7 & 4.1 & 6.2 \\
Dutch & \textbf{2.8} & 3.5 & 3.5 & 3.7 & 4.7 \\
Urdu & 20.8 & 31.9 & 25.2 & \textbf{19.7} & 23.0 \\
Norwegian & \textbf{3.9} & 5.2 & 5.0 & 5.5 & 6.7 \\
Swedish & \textbf{3.1} & 5.0 & 4.6 & 5.2 & 7.7 \\
Danish & \textbf{3.5} & 5.3 & 5.7 & 6.5 & 7.9 \\
Hebrew & \textbf{12.5} & 16.6 & 15.6 & 19.4 & 20.2 \\
Finnish & \textbf{2.4} & 4.5 & 3.4 & 3.8 & 5.2 \\
Polish & \textbf{1.9} & 3.1 & 2.7 & 2.8 & 4.9 \\
Icelandic & \textbf{3.6} & 8.9 & 4.7 & 10.8 & 6.8 \\
Czech & \textbf{2.6} & 4.5 & 3.8 & 4.7 & 8.0 \\
Filipino & \textbf{5.1} & 7.1 & 7.6 & 7.3 & 8.5 \\
Persian & 12.0 & 12.1 & \textbf{8.9} & 9.9 & 10.0 \\
Greek & \textbf{4.7} & 8.1 & 5.4 & 6.5 & 7.6 \\
Afrikaans & \textbf{10.6} & 13.7 & 12.7 & 17.9 & 18.6 \\
Asturian & \textbf{15.8} & 25.9 & 23.7 & 23.8 & 48.3 \\
Belarusian & \textbf{6.7} & 12.2 & \textbf{6.7} & 10.2 & 10.9 \\
Bulgarian & 6.2 & 10.7 & \textbf{5.3} & 7.0 & 7.9 \\
Bengali & \textbf{16.2} & 19.8 & 21.9 & 24.1 & 21.9 \\
Bosnian & \textbf{5.4} & 9.5 & 6.0 & 13.9 & 11.2 \\
Catalan & 2.8 & 6.3 & \textbf{2.7} & \textbf{2.7} & 6.4 \\
Cebuano & \textbf{10.5} & 16.6 & 13.0 & 15.1 & 12.8 \\
Estonian & 6.7 & 16.6 & \textbf{4.9} & 7.6 & 7.3 \\
Galician & 5.0 & 8.6 & \textbf{4.9} & 6.8 & 14.6 \\
Gujarati & \textbf{13.9} & 18.4 & 14.8 & 26.9 & 16.3 \\
Croatian & 5.4 & 9.0 & \textbf{5.1} & 16.4 & 9.3 \\
Hungarian & \textbf{4.9} & 10.6 & 5.5 & 7.4 & 11.3 \\
Javanese & \textbf{11.8} & 18.3 & 14.1 & 24.7 & 15.7 \\
Kazakh & 6.3 & 16.6 & \textbf{6.2} & 11.5 & 12.7 \\
Kannada & \textbf{16.0} & 23.8 & 16.3 & 28.1 & 16.3 \\
Kyrgyz & 10.0 & 19.7 & \textbf{8.3} & 20.7 & 16.3 \\
Latvian & 6.7 & 17.8 & \textbf{3.7} & 6.3 & 6.8 \\
Macedonian & 4.1 & 7.9 & \textbf{4.0} & 6.2 & 7.8 \\
Malayalam & 18.8 & 27.0 & \textbf{18.3} & 33.5 & 20.3 \\
Marathi & 16.3 & 23.6 & \textbf{15.3} & 26.3 & 16.0 \\
Punjabi & \textbf{13.7} & 24.6 & 14.4 & 36.4 & 17.4 \\
Romanian & \textbf{3.2} & 6.1 & 3.4 & 4.5 & 5.9 \\
Slovak & 3.3 & 5.5 & \textbf{2.8} & 3.6 & 6.5 \\
Slovenian & \textbf{6.1} & 14.3 & 6.3 & 8.8 & 10.0 \\
Swahili & \textbf{9.4} & 17.5 & 9.9 & 16.3 & 10.7 \\
Tajik & \textbf{10.0} & 41.1 & 20.4 & 20.2 & 53.1 \\
Azerbaijani & 7.2 & 13.0 & \textbf{5.8} & 10.6 & 13.2 \\
Ukrainian & \textbf{3.2} & 5.4 & 3.4 & 4.3 & 5.1 \\
\midrule
Average & \textbf{6.6} & 10.8 & 7.3 & 10.4 & 10.5 \\
\bottomrule
\end{tabular}
\label{tab:multilingual_asr_detail_table}
\end{table}

\begin{table}[htbp]
\centering
\small
\caption{Multilingual translation performance on the FLEURS en2xx and zh2xx test sets. Results are reported using BLEU (↑). Compared with competing models, Qwen3.5-Omni-Plus outperforms them on the majority of language pairs. The best results are highlighted in bold.}
\setlength\tabcolsep{2pt}
\begin{tabular}{lcccccccc}
\toprule
& \multicolumn{4}{c}{\textbf{en2xx (English → Other Languages)}} & \multicolumn{4}{c}{\textbf{zh2xx (Chinese → Other Languages)}} \\
\cmidrule(lr){2-5} \cmidrule(lr){6-9}
\textbf{Language} 
& \begin{tabular}[c]{@{}c@{}}\textbf{Qwen3.5-}\\ \textbf{Omni-Plus}\end{tabular} 
& \begin{tabular}[c]{@{}c@{}}\textbf{Qwen3.5-}\\ \textbf{Omni-Flash}\end{tabular} 
& \begin{tabular}[c]{@{}c@{}}\textbf{Gemin-3.1-}\\ \textbf{Pro}\end{tabular} 
& \begin{tabular}[c]{@{}c@{}}\textbf{Gemini-3-}\\ \textbf{Flash}\end{tabular}
& \begin{tabular}[c]{@{}c@{}}\textbf{Qwen3.5-}\\ \textbf{Omni-Plus}\end{tabular} 
& \begin{tabular}[c]{@{}c@{}}\textbf{Qwen3.5-}\\ \textbf{Omni-Flash}\end{tabular} 
& \begin{tabular}[c]{@{}c@{}}\textbf{Gemin-3.1-}\\ \textbf{Pro}\end{tabular}
& \begin{tabular}[c]{@{}c@{}}\textbf{Gemini-3-}\\ \textbf{Flash}\end{tabular} \\
\midrule
Chinese & \textbf{47.8} & 46.6 & 47.4 & 46.3 & -- & -- & -- & -- \\
English & -- & -- & -- & -- & \textbf{32.2} & 31.2 & 30.1 & 29.5 \\
Cantonese & \textbf{40.1} & 37.3 & 25.5 & 26.7 & \textbf{36.7} & 35.9 & 23.7 & 24.0 \\
Arabic & \textbf{31.1} & 28.2 & 27.0 & 28.5 & \textbf{16.1} & 13.9 & 14.2 & 14.4 \\
German & \textbf{43.2} & 39.6 & 41.8 & 40.9 & \textbf{23.2} & 20.8 & 22.0 & 21.4 \\
French & \textbf{50.9} & 48.8 & 48.0 & 47.8 & \textbf{30.7} & 28.8 & 29.4 & 29.2 \\
Spanish & \textbf{29.1} & 28.9 & 28.3 & 28.8 & \textbf{22.2} & 20.4 & 20.8 & 20.6 \\
Portuguese & \textbf{51.2} & 48.6 & 47.3 & 47.2 & \textbf{28.5} & 26.8 & 25.7 & 25.6 \\
Indonesian & \textbf{45.3} & 43.7 & 42.1 & 41.7 & \textbf{28.8} & 26.9 & 25.4 & 25.2 \\
Italian & \textbf{32.7} & 30.7 & 31.9 & 30.9 & \textbf{23.1} & 21.1 & 21.9 & 21.3 \\
Korean & \textbf{33.9} & 31.8 & 30.8 & 31.7 & \textbf{25.1} & 23.4 & 21.9 & 22.8 \\
Russian & \textbf{33.8} & 31.8 & 33.1 & 33.2 & \textbf{21.5} & 18.9 & 20.0 & 19.9 \\
Thai & \textbf{65.4} & 62.9 & 64.8 & 64.4 & \textbf{58.0} & 55.5 & 57.2 & 56.4 \\
Vietnamese & \textbf{43.0} & 41.8 & 41.1 & 40.2 & \textbf{31.6} & 30.5 & 28.1 & 28.4 \\
Japanese & \textbf{53.2} & 50.6 & 51.3 & 50.8 & \textbf{45.6} & 41.6 & 43.0 & 42.0 \\
Turkish & \textbf{30.4} & 27.6 & 29.3 & 29.0 & \textbf{16.8} & 14.6 & 15.9 & 16.0 \\
Hindi & \textbf{33.1} & 29.1 & 28.2 & 29.2 & \textbf{19.1} & 14.3 & 17.6 & 17.5 \\
Malay & \textbf{39.6} & 37.2 & 35.9 & 36.0 & \textbf{24.1} & 21.7 & 21.0 & 20.5 \\
Dutch & \textbf{30.1} & 28.2 & 28.1 & 28.8 & \textbf{21.0} & 18.8 & 19.3 & 19.1 \\
Urdu & \textbf{25.0} & 22.1 & 23.0 & 23.0 & \textbf{15.5} & 8.6 & 14.9 & 14.7 \\
Norwegian & \textbf{35.3} & 32.8 & 33.1 & 33.7 & \textbf{20.3} & 17.8 & 18.4 & 18.9 \\
Swedish & \textbf{47.5} & 44.1 & 45.7 & 45.8 & \textbf{25.4} & 23.0 & 23.7 & 24.1 \\
Danish & \textbf{48.4} & 45.2 & 45.7 & 45.2 & \textbf{25.7} & 22.8 & 23.5 & 23.2 \\
Hebrew & 36.4 & 29.9 & \textbf{36.5} & 35.4 & \textbf{18.2} & 14.5 & 17.7 & 17.4 \\
Finnish & 30.1 & 26.0 & 32.1 & \textbf{32.3} & 18.1 & 15.3 & \textbf{18.6} & 17.7 \\
Polish & \textbf{25.2} & 22.5 & 24.9 & 23.4 & \textbf{17.5} & 15.0 & 15.6 & 15.5 \\
Icelandic & 28.5 & 27.2 & \textbf{29.6} & 28.2 & \textbf{16.2} & 13.5 & 16.0 & 15.6 \\
Czech & \textbf{35.9} & 32.5 & 33.3 & 33.7 & \textbf{20.4} & 18.1 & 19.4 & 19.0 \\
Filipino & \textbf{35.0} & 32.0 & 32.1 & 33.1 & \textbf{22.3} & 19.0 & 20.7 & 20.7 \\
Persian & \textbf{30.7} & 27.3 & 25.1 & 25.9 & \textbf{19.5} & 16.4 & 15.9 & 15.9 \\
Greek & \textbf{30.0} & 27.8 & \textbf{30.0} & 29.4 & \textbf{18.4} & 15.9 & 17.6 & 17.5 \\
Asturian & \textbf{32.4} & 27.9 & 31.5 & 30.4 & \textbf{20.4} & 16.2 & 18.6 & 18.1 \\
Belarusian & 16.4 & 14.7 & 16.4 & \textbf{16.5} & \textbf{12.6} & 10.8 & 12.1 & 12.2 \\
Bulgarian & \textbf{45.0} & 40.7 & 41.7 & 42.5 & \textbf{25.6} & 23.0 & 24.3 & 24.0 \\
Bengali & \textbf{18.6} & 15.7 & 14.3 & 15.0 & \textbf{10.6} & 9.2 & 9.0 & 9.3 \\
Bosnian & \textbf{37.5} & 34.0 & 36.3 & 35.2 & \textbf{21.4} & 18.6 & 19.9 & 19.4 \\
Catalan & \textbf{43.9} & 41.5 & 42.7 & 42.9 & \textbf{26.6} & 17.2 & 25.0 & 25.1 \\
Cebuano & 28.5 & 12.7 & 28.5 & \textbf{29.2} & \textbf{19.0} & 5.6 & 17.5 & 17.7 \\
Estonian & 30.8 & 26.3 & \textbf{31.4} & 30.6 & \textbf{18.9} & 13.6 & 17.5 & 17.2 \\
Galician & \textbf{37.4} & 35.4 & 36.6 & 35.9 & \textbf{23.9} & 22.0 & 22.7 & 22.3 \\
Gujarati & \textbf{23.8} & 20.9 & 21.0 & 21.5 & \textbf{14.3} & 10.8 & 12.3 & 12.3 \\
Croatian & 33.3 & 30.7 & \textbf{33.4} & 32.6 & \textbf{21.3} & 18.5 & 19.2 & 18.6 \\
Hungarian & \textbf{29.5} & 24.9 & 28.1 & 27.6 & \textbf{18.8} & 15.8 & 17.3 & 16.7 \\
Javanese & \textbf{26.8} & 24.4 & 16.4 & 22.1 & \textbf{16.5} & 14.8 & 9.0 & 13.0 \\
Kazakh & \textbf{24.9} & 21.1 & 20.6 & 22.5 & \textbf{15.0} & 12.4 & 12.9 & 13.1 \\
Kannada & \textbf{20.0} & 17.0 & 16.1 & 17.2 & \textbf{11.7} & 6.9 & 9.5 & 10.0 \\
Kyrgyz & \textbf{15.3} & 12.6 & 15.1 & 14.7 & \textbf{10.4} & 7.8 & 10.0 & 9.5 \\
Latvian & \textbf{36.1} & 31.0 & 35.4 & 35.3 & \textbf{21.9} & 17.8 & 20.1 & 19.5 \\
Macedonian & 38.1 & 34.0 & \textbf{38.5} & 38.2 & \textbf{22.3} & 20.1 & 22.0 & 21.6 \\
Malayalam & \textbf{19.3} & 11.1 & 16.1 & 16.1 & \textbf{10.4} & 5.2 & 9.9 & 9.4 \\
Marathi & \textbf{17.7} & 11.6 & 15.9 & 16.2 & \textbf{11.3} & 8.1 & 9.4 & 10.1 \\
Punjabi & \textbf{26.1} & 23.1 & 24.1 & 24.6 & \textbf{15.7} & 8.9 & 14.3 & 14.1 \\
Romanian & \textbf{42.0} & 39.9 & 41.4 & \textbf{42.0} & \textbf{25.6} & 22.8 & 23.6 & 23.4 \\
Slovak & \textbf{35.3} & 31.4 & 34.8 & 34.6 & \textbf{19.6} & 16.4 & 19.2 & 18.5 \\
Slovenian & 32.8 & 28.5 & \textbf{33.5} & 32.9 & 20.1 & 17.7 & \textbf{20.9} & 20.2 \\
Swahili & \textbf{36.3} & 30.9 & 32.1 & 32.2 & \textbf{20.4} & 9.3 & 18.5 & 18.3 \\
Tajik & \textbf{23.8} & 18.3 & 22.1 & 22.9 & \textbf{14.6} & 10.9 & 14.3 & 14.2 \\
Azerbaijani & 13.7 & 9.8 & \textbf{15.2} & 14.7 & \textbf{11.5} & 9.7 & 11.3 & 10.8 \\
Ukrainian & \textbf{31.7} & 29.2 & 29.8 & 30.1 & \textbf{19.5} & 15.0 & 17.0 & 17.6 \\
\midrule
Average & \textbf{33.8} & 30.4 & 31.8 & 31.8 & \textbf{21.4} & 18.1 & 19.6 & 19.5 \\
\bottomrule
\end{tabular}
\label{tab:multilingual_speech_translation_2xx_table}
\end{table}

\begin{table}[htbp]
\centering
\small
\caption{Multilingual translation performance on the FLEURS xx2en and xx2zh test sets. Results are reported using BLEU (↑). The best results are highlighted in bold.}
\setlength\tabcolsep{2pt}
\begin{tabular}{lcccccccc}
\toprule
& \multicolumn{4}{c}{\textbf{xx2en (Other Languages → English)}} & \multicolumn{4}{c}{\textbf{xx2zh (Other Languages → Chinese)}} \\
\cmidrule(lr){2-5} \cmidrule(lr){6-9}
\textbf{Language} 
& \begin{tabular}[c]{@{}c@{}}\textbf{Qwen3.5-}\\ \textbf{Omni-Plus}\end{tabular} 
& \begin{tabular}[c]{@{}c@{}}\textbf{Qwen3.5-}\\ \textbf{Omni-Flash}\end{tabular} 
& \begin{tabular}[c]{@{}c@{}}\textbf{Gemin-3.1-}\\ \textbf{Pro}\end{tabular} 
& \begin{tabular}[c]{@{}c@{}}\textbf{Gemini-3-}\\ \textbf{Flash}\end{tabular}
& \begin{tabular}[c]{@{}c@{}}\textbf{Qwen3.5-}\\ \textbf{Omni-Plus}\end{tabular} 
& \begin{tabular}[c]{@{}c@{}}\textbf{Qwen3.5-}\\ \textbf{Omni-Flash}\end{tabular} 
& \begin{tabular}[c]{@{}c@{}}\textbf{Gemin-3.1-}\\ \textbf{Pro}\end{tabular}
& \begin{tabular}[c]{@{}c@{}}\textbf{Gemini-3-}\\ \textbf{Flash}\end{tabular} \\
\midrule
Chinese & \textbf{32.2} & 31.2 & 30.1 & 29.5 & -- & -- & -- & -- \\
English & -- & -- & -- & -- & \textbf{47.8} & 46.6 & 47.4 & 46.3 \\
Cantonese & \textbf{30.3} & 29.9 & 27.7 & 26.3 & 36.8 & \textbf{37.5} & 21.2 & 22.4 \\
Arabic & \textbf{42.9} & 40.0 & 42.1 & 42.3 & 40.2 & 37.3 & \textbf{40.5} & \textbf{40.5} \\
German & \textbf{44.6} & 44.1 & 43.9 & 43.9 & \textbf{43.3} & 42.8 & 42.1 & 41.8 \\
French & \textbf{43.5} & 42.0 & 41.6 & 41.3 & \textbf{41.6} & 40.3 & \textbf{41.6} & 41.4 \\
Spanish & \textbf{32.3} & 31.3 & 30.3 & 30.4 & \textbf{38.8} & 38.5 & 38.4 & 38.2 \\
Portuguese & \textbf{49.4} & 48.2 & 47.7 & 47.5 & \textbf{43.6} & 41.8 & 42.5 & 42.4 \\
Indonesian & \textbf{45.7} & 43.1 & 45.1 & 44.9 & \textbf{43.5} & 41.4 & 42.8 & 42.9 \\
Italian & \textbf{34.4} & 31.9 & 30.8 & 31.0 & \textbf{40.8} & 39.5 & 39.7 & 39.4 \\
Korean & \textbf{34.1} & 32.4 & 32.0 & 32.1 & \textbf{39.9} & 37.5 & 37.0 & 37.2 \\
Russian & \textbf{38.6} & 37.2 & 36.7 & 36.5 & \textbf{41.7} & 39.5 & 41.1 & 40.4 \\
Thai & 34.1 & 32.4 & \textbf{34.2} & 33.0 & \textbf{40.2} & 37.9 & 40.0 & 39.5 \\
Vietnamese & \textbf{36.4} & 34.9 & 36.1 & 35.4 & 38.7 & 36.3 & \textbf{39.2} & 38.9 \\
Japanese & \textbf{30.4} & 29.2 & 29.5 & 29.4 & \textbf{38.0} & 35.7 & 35.6 & 34.8 \\
Turkish & \textbf{40.3} & 39.1 & 39.5 & 38.9 & \textbf{41.8} & 40.3 & 40.6 & 41.0 \\
Hindi & 38.8 & 36.2 & \textbf{39.3} & 39.2 & \textbf{38.9} & 36.9 & 38.0 & 38.4 \\
Malay & 42.9 & 41.1 & \textbf{44.8} & 42.4 & 41.0 & 39.5 & \textbf{42.4} & 41.5 \\
Dutch & \textbf{33.3} & 32.2 & 31.2 & 30.4 & \textbf{40.1} & 38.5 & 39.4 & 39.4 \\
Urdu & \textbf{35.5} & 31.5 & 32.9 & 32.6 & \textbf{37.2} & 33.8 & 36.9 & 36.5 \\
Norwegian & \textbf{43.5} & 42.1 & 42.6 & 41.4 & \textbf{42.2} & 39.6 & 40.9 & 40.8 \\
Swedish & \textbf{47.2} & 45.2 & 46.6 & 44.7 & \textbf{42.9} & 40.7 & 41.5 & 41.6 \\
Danish & \textbf{45.4} & 44.0 & 44.7 & 42.7 & \textbf{43.4} & 41.1 & 41.7 & 40.6 \\
Hebrew & 39.7 & 36.4 & \textbf{42.5} & 39.9 & 36.7 & 34.1 & \textbf{40.3} & 37.8 \\
Finnish & \textbf{36.9} & 35.0 & 36.7 & 35.5 & 40.8 & 38.7 & \textbf{41.0} & 40.6 \\
Polish & \textbf{32.1} & 30.5 & 30.4 & 29.7 & \textbf{38.4} & 36.0 & 37.9 & 37.4 \\
Icelandic & 31.5 & 27.5 & 35.1 & \textbf{35.8} & \textbf{38.2} & 31.8 & 37.9 & 37.5 \\
Czech & \textbf{42.1} & 39.3 & 40.1 & 39.5 & \textbf{40.6} & 39.4 & \textbf{40.6} & 40.0 \\
Filipino & 42.7 & 40.9 & \textbf{45.0} & 44.3 & 41.0 & 38.0 & \textbf{42.6} & 41.5 \\
Persian & \textbf{40.2} & 36.8 & 38.0 & 38.1 & 40.2 & 37.0 & \textbf{41.1} & \textbf{41.1} \\
Greek & \textbf{36.0} & 32.5 & 35.3 & 35.4 & 38.0 & 32.8 & \textbf{39.0} & 38.4 \\
Asturian & 37.0 & 35.1 & \textbf{37.2} & 35.5 & 37.7 & 34.0 & \textbf{38.1} & 36.7 \\
Belarusian & \textbf{23.1} & 19.9 & 20.6 & 19.9 & 33.3 & 31.2 & \textbf{33.7} & \textbf{33.7} \\
Bulgarian & 39.6 & 36.0 & \textbf{41.3} & 40.0 & 40.9 & 36.6 & \textbf{41.0} & 40.4 \\
Bengali & 32.0 & 27.6 & \textbf{34.9} & 34.3 & 35.8 & 32.4 & \textbf{38.1} & 37.5 \\
Bosnian & \textbf{43.1} & 40.8 & 42.7 & 42.4 & 41.5 & 39.1 & \textbf{42.3} & 41.4 \\
Catalan & \textbf{46.6} & 42.3 & 46.2 & 45.5 & 42.2 & 38.9 & \textbf{42.6} & 41.5 \\
Cebuano & 37.3 & 26.3 & \textbf{38.9} & 38.2 & 34.0 & 26.5 & \textbf{36.7} & 36.6 \\
Estonian & 35.7 & 28.3 & \textbf{40.1} & 38.3 & 38.0 & 32.2 & \textbf{41.7} & 41.2 \\
Galician & \textbf{40.8} & 38.6 & 39.6 & 38.6 & 40.9 & 39.1 & \textbf{41.6} & 40.9 \\
Gujarati & 33.4 & 28.3 & \textbf{40.3} & 39.7 & 35.8 & 31.4 & \textbf{39.7} & 39.0 \\
Croatian & \textbf{39.5} & 36.3 & 38.0 & 36.9 & 40.0 & 37.7 & \textbf{40.2} & 39.6 \\
Hungarian & \textbf{35.5} & 29.6 & 35.3 & 33.3 & 39.1 & 33.9 & \textbf{40.0} & 37.4 \\
Javanese & 35.9 & 28.4 & \textbf{38.4} & 36.5 & 34.9 & 28.3 & \textbf{36.7} & 36.3 \\
Kazakh & 34.4 & 26.9 & \textbf{35.7} & 35.5 & 37.1 & 31.4 & \textbf{39.3} & 38.8 \\
Kannada & 26.4 & 19.8 & 33.1 & \textbf{33.6} & 32.3 & 26.1 & \textbf{38.0} & 37.8 \\
Kyrgyz & 22.2 & 17.1 & \textbf{24.8} & 23.7 & 29.9 & 24.8 & \textbf{33.7} & 32.8 \\
Latvian & 33.7 & 25.2 & \textbf{38.0} & 37.9 & 37.1 & 30.0 & \textbf{41.4} & 40.6 \\
Macedonian & \textbf{43.3} & 39.6 & 43.1 & 41.9 & 41.6 & 38.0 & \textbf{42.1} & 41.8 \\
Malayalam & 31.2 & 25.7 & \textbf{34.9} & 34.2 & 36.1 & 31.5 & \textbf{38.4} & 38.0 \\
Marathi & 33.5 & 25.7 & \textbf{36.7} & 35.5 & 34.6 & 29.4 & \textbf{38.8} & 37.9 \\
Punjabi & 33.0 & 26.9 & \textbf{38.3} & 36.5 & 35.1 & 29.9 & \textbf{37.4} & 36.4 \\
Romanian & \textbf{43.5} & 39.7 & 42.3 & 41.5 & 42.0 & 38.7 & \textbf{42.4} & 41.3 \\
Slovak & 39.7 & 38.4 & \textbf{39.9} & 38.8 & 39.2 & 38.1 & \textbf{40.2} & 39.3 \\
Slovenian & 31.7 & 26.5 & \textbf{34.7} & 33.2 & 34.5 & 30.1 & \textbf{38.4} & 37.3 \\
Swahili & 35.0 & 27.4 & \textbf{42.6} & 40.5 & 33.9 & 26.8 & \textbf{39.2} & 37.9 \\
Tajik & 33.9 & 29.0 & \textbf{34.5} & 33.3 & 36.7 & 32.7 & \textbf{38.9} & 38.1 \\
Azerbaijani & \textbf{25.0} & 22.0 & 23.7 & 23.3 & 33.4 & 30.5 & \textbf{33.5} & 33.3 \\
Ukrainian & \textbf{42.0} & 40.1 & 41.7 & 41.4 & \textbf{41.7} & 39.7 & 41.6 & 40.7 \\
\midrule
Average & 37.0 & 33.5 & \textbf{37.4} & 36.6 & 38.9 & 35.7 & \textbf{39.4} & 38.9 \\
\bottomrule
\end{tabular}
\label{tab:multilingual_speech_translation_xx2_table}
\end{table}

%% file: biblio.bib
@string {CVPR = "Conference on Computer Vision and Pattern Recognition (CVPR)"}

@string {ICCV = "International Conference on Computer Vision (ICCV)"}

@string {ECCV = "European Conference on Computer Vision (ECCV)"}

@string {ICML = "International Conference on Machine Learning (ICML)"}

@string {ICLR = "International Conference on Learning Representations (ICLR)"}

@string {ACL = "Annual Meeting of the Association for Computational Linguistics (ACL)"}

@string {EMNLP = "Annual Conference on Empirical Methods in Natural Language Processing (EMNLP)"}

@string {CHI = "ACM Conference on Human Factors in Computing Systems (CHI)"}

@inproceedings{rafailov2024direct,
  author       = {Rafael Rafailov and
                  Archit Sharma and
                  Eric Mitchell and
                  Christopher D. Manning and
                  Stefano Ermon and
                  Chelsea Finn},
  title        = {Direct Preference Optimization: Your Language Model is Secretly a
                  Reward Model},
  booktitle    = {NeurIPS},
  year         = {2023}
}

@misc{chatml,
  title = {{ChatML}},
  author = {{OpenAI}},
  url = {https://github.com/openai/openai-python/blob/e389823ba013a24b4c32ce38fa0bd87e6bccae94/chatml.md},
  year = {2022}
}

@misc{claude,
  title = {Introducing {Claude}},
  author = {Anthropic},
  institution = {Anthropic},
  url = {https://www.anthropic.com/index/introducing-claude},
  year={2023}
}

@techreport{claude2,
  title = {Claude 2},
  author = {Anthropic},
  institution = {Anthropic},
  url = {https://www-files.anthropic.com/production/images/Model-Card-Claude-2.pdf},
  year={2023}
}

@article{gpt4,
  title={{GPT4} technical report},
  author={{OpenAI}},
  journal={CoRR},
  volume={abs/2303.08774},
  year={2023}
}

@article{mmluredux,
  title={Are We Done with MMLU?},
  author={Gema, Aryo Pradipta and Leang, Joshua Ong Jun and Hong, Giwon and Devoto, Alessio and Mancino, Alberto Carlo Maria and Saxena, Rohit and He, Xuanli and Zhao, Yu and Du, Xiaotang and Madani, Mohammad Reza Ghasemi and others},
  journal={CoRR},
  volume={abs/2406.04127},
  year={2024}
}

@inproceedings{ceval,
  author       = {Yuzhen Huang and
                  Yuzhuo Bai and
                  Zhihao Zhu and
                  Junlei Zhang and
                  Jinghan Zhang and
                  Tangjun Su and
                  Junteng Liu and
                  Chuancheng Lv and
                  Yikai Zhang and
                  Jiayi Lei and
                  Yao Fu and
                  Maosong Sun and
                  Junxian He},
  title        = {{C-Eval}: A Multi-Level Multi-Discipline Chinese Evaluation Suite
                  for Foundation Models},
  booktitle    = {NeurIPS},
  year         = {2023}
}

@article{qwenvl,
  author       = {Jinze Bai and
                  Shuai Bai and
                  Shusheng Yang and
                  Shijie Wang and
                  Sinan Tan and
                  Peng Wang and
                  Junyang Lin and
                  Chang Zhou and
                  Jingren Zhou},
  title        = {{Qwen-VL}: A Frontier Large Vision-Language Model with Versatile Abilities},
  journal      = {CoRR},
  volume       = {abs/2308.12966},
  year         = {2023}
}

@article{livecodebench,
  author       = {Naman Jain and
                  King Han and
                  Alex Gu and
                  Wen{-}Ding Li and
                  Fanjia Yan and
                  Tianjun Zhang and
                  Sida Wang and
                  Armando Solar{-}Lezama and
                  Koushik Sen and
                  Ion Stoica},
  title        = {{LiveCodeBench}: Holistic and Contamination Free Evaluation of Large
                  Language Models for Code},
  journal      = {CoRR},
  volume       = {abs/2403.07974},
  year         = {2024}
}

@article{pro,
  title={Preference ranking optimization for human alignment},
  author={Song, Feifan and Yu, Bowen and Li, Minghao and Yu, Haiyang and Huang, Fei and Li, Yongbin and Wang, Houfeng},
  journal={CoRR},
  volume={abs/2306.17492},
  year={2023}
}

@article{qwen,
author       = {Jinze Bai and
                  Shuai Bai and
                  Yunfei Chu and
                  Zeyu Cui and
                  Kai Dang and
                  Xiaodong Deng and
                  Yang Fan and
                  Wenbin Ge and
                  Yu Han and
                  Fei Huang and
                  Binyuan Hui and
                  Luo Ji and
                  Mei Li and
                  Junyang Lin and
                  Runji Lin and
                  Dayiheng Liu and
                  Gao Liu and
                  Chengqiang Lu and
                  Keming Lu and
                  Jianxin Ma and
                  Rui Men and
                  Xingzhang Ren and
                  Xuancheng Ren and
                  Chuanqi Tan and
                  Sinan Tan and
                  Jianhong Tu and
                  Peng Wang and
                  Shijie Wang and
                  Wei Wang and
                  Shengguang Wu and
                  Benfeng Xu and
                  Jin Xu and
                  An Yang and
                  Hao Yang and
                  Jian Yang and
                  Shusheng Yang and
                  Yang Yao and
                  Bowen Yu and
                  Hongyi Yuan and
                  Zheng Yuan and
                  Jianwei Zhang and
                  Xingxuan Zhang and
                  Yichang Zhang and
                  Zhenru Zhang and
                  Chang Zhou and
                  Jingren Zhou and
                  Xiaohuan Zhou and
                  Tianhang Zhu},
  title        = {Qwen Technical Report},
  journal      = {CoRR},
  volume       = {abs/2309.16609},
  year         = {2023}
}

@techreport{claude3,
  title={The {Claude} 3 model family: {Opus}, {Sonnet}, {Haiku}},
  author={Anthropic},
  institution={{Anthropic, AI}},
  url={https://www-cdn.anthropic.com/de8ba9b01c9ab7cbabf5c33b80b7bbc618857627/Model\_Card\_Claude\_3.pdf},
  year={2024}
}

@misc{gpt4o,
  title = {Hello {GPT-4o}},
  author = {{OpenAI}},
  url = {https://openai.com/index/hello-gpt-4o/},
  year = {2024}
}

@article{llama3,
  author       = {Abhimanyu Dubey and
                  Abhinav Jauhri and
                  Abhinav Pandey and
                  Abhishek Kadian and
                  Ahmad Al{-}Dahle and
                  Aiesha Letman and
                  Akhil Mathur and
                  Alan Schelten and
                  Amy Yang and
                  Angela Fan and
                  Anirudh Goyal and
                  Anthony Hartshorn and
                  Aobo Yang and
                  Archi Mitra and
                  Archie Sravankumar and
                  Artem Korenev and
                  Arthur Hinsvark and
                  Arun Rao and
                  Aston Zhang and
                  Aur{\'{e}}lien Rodriguez and
                  Austen Gregerson and
                  Ava Spataru and
                  Baptiste Rozi{\`{e}}re and
                  Bethany Biron and
                  Binh Tang and
                  Bobbie Chern and
                  Charlotte Caucheteux and
                  Chaya Nayak and
                  Chloe Bi and
                  Chris Marra and
                  Chris McConnell and
                  Christian Keller and
                  Christophe Touret and
                  Chunyang Wu and
                  Corinne Wong and
                  Cristian Canton Ferrer and
                  Cyrus Nikolaidis and
                  Damien Allonsius and
                  Daniel Song and
                  Danielle Pintz and
                  Danny Livshits and
                  David Esiobu and
                  Dhruv Choudhary and
                  Dhruv Mahajan and
                  Diego Garcia{-}Olano and
                  Diego Perino and
                  Dieuwke Hupkes and
                  Egor Lakomkin and
                  Ehab AlBadawy and
                  Elina Lobanova and
                  Emily Dinan and
                  Eric Michael Smith and
                  Filip Radenovic and
                  Frank Zhang and
                  Gabriel Synnaeve and
                  Gabrielle Lee and
                  Georgia Lewis Anderson and
                  Graeme Nail and
                  Gr{\'{e}}goire Mialon and
                  Guan Pang and
                  Guillem Cucurell and
                  Hailey Nguyen and
                  Hannah Korevaar and
                  Hu Xu and
                  Hugo Touvron and
                  Iliyan Zarov and
                  Imanol Arrieta Ibarra and
                  Isabel M. Kloumann and
                  Ishan Misra and
                  Ivan Evtimov and
                  Jade Copet and
                  Jaewon Lee and
                  Jan Geffert and
                  Jana Vranes and
                  Jason Park and
                  Jay Mahadeokar and
                  Jeet Shah and
                  Jelmer van der Linde and
                  Jennifer Billock and
                  Jenny Hong and
                  Jenya Lee and
                  Jeremy Fu and
                  Jianfeng Chi and
                  Jianyu Huang and
                  Jiawen Liu and
                  Jie Wang and
                  Jiecao Yu and
                  Joanna Bitton and
                  Joe Spisak and
                  Jongsoo Park and
                  Joseph Rocca and
                  Joshua Johnstun and
                  Joshua Saxe and
                  Junteng Jia and
                  Kalyan Vasuden Alwala and
                  Kartikeya Upasani and
                  Kate Plawiak and
                  Ke Li and
                  Kenneth Heafield and
                  Kevin Stone and
                  et al.},
  title        = {The {Llama} 3 Herd of Models},
  journal      = {CoRR},
  volume       = {abs/2407.21783},
  year         = {2024}
}

@techreport{gemini,
  title = {Gemini 1.5: Unlocking multimodal understanding across millions of tokens of context},
  author = {{Gemini Team}},
  institution = {Google},
  url = {https://storage.googleapis.com/deepmind-media/gemini/gemini\_v1\_5\_report.pdf},
  year = {2024}
}

@article{qwenaudio,
  author       = {Yunfei Chu and
                  Jin Xu and
                  Xiaohuan Zhou and
                  Qian Yang and
                  Shiliang Zhang and
                  Zhijie Yan and
                  Chang Zhou and
                  Jingren Zhou},
  title        = {{Qwen-Audio}: Advancing Universal Audio Understanding via Unified Large-Scale
                  Audio-Language Models},
  journal      = {CoRR},
  volume       = {abs/2311.07919},
  year         = {2023}
}

@article{gpqa,
  author       = {David Rein and
                  Betty Li Hou and
                  Asa Cooper Stickland and
                  Jackson Petty and
                  Richard Yuanzhe Pang and
                  Julien Dirani and
                  Julian Michael and
                  Samuel R. Bowman},
  title        = {{GPQA}: A Graduate-Level {Google}-Proof {Q}{\&}{A} Benchmark},
  journal      = {CoRR},
  volume       = {abs/2311.12022},
  year         = {2023}
}

@article{ifeval,
  author       = {Jeffrey Zhou and
                  Tianjian Lu and
                  Swaroop Mishra and
                  Siddhartha Brahma and
                  Sujoy Basu and
                  Yi Luan and
                  Denny Zhou and
                  Le Hou},
  title        = {Instruction-Following Evaluation for Large Language Models},
  journal      = {CoRR},
  volume       = {abs/2311.07911},
  year         = {2023}
}

@article{mmlupro,
  author       = {Yubo Wang and
                  Xueguang Ma and
                  Ge Zhang and
                  Yuansheng Ni and
                  Abhranil Chandra and
                  Shiguang Guo and
                  Weiming Ren and
                  Aaran Arulraj and
                  Xuan He and
                  Ziyan Jiang and
                  Tianle Li and
                  Max Ku and
                  Kai Wang and
                  Alex Zhuang and
                  Rongqi Fan and
                  Xiang Yue and
                  Wenhu Chen},
  title        = {{MMLU-Pro}: {A} More Robust and Challenging Multi-Task Language Understanding
                  Benchmark},
  journal      = {CoRR},
  volume       = {abs/2406.01574},
  year         = {2024}
}

@article{blip2,
  title={Blip-2: Bootstrapping language-image pre-training with frozen image encoders and large language models},
  author={Li, Junnan and Li, Dongxu and Savarese, Silvio and Hoi, Steven},
  journal={arXiv:2301.12597},
  year={2023}
}

@article{fu2024video,
  title={Video-MME: The First-Ever Comprehensive Evaluation Benchmark of Multi-modal LLMs in Video Analysis},
  author={Fu, Chaoyou and Dai, Yuhan and Luo, Yondong and Li, Lei and Ren, Shuhuai and Zhang, Renrui and Wang, Zihan and Zhou, Chenyu and Shen, Yunhang and Zhang, Mengdan and others},
  journal={arXiv:2405.21075},
  year={2024}
}

@inproceedings{li2024mvbench,
  title={Mvbench: A comprehensive multi-modal video understanding benchmark},
  author={Li, Kunchang and Wang, Yali and He, Yinan and Li, Yizhuo and Wang, Yi and Liu, Yi and Wang, Zun and Xu, Jilan and Chen, Guo and Luo, Ping and others},
  booktitle={CVPR},
  year={2024}
}

@article{chen2024we,
  title={Are We on the Right Way for Evaluating Large Vision-Language Models?},
  author={Chen, Lin and Li, Jinsong and Dong, Xiaoyi and Zhang, Pan and Zang, Yuhang and Chen, Zehui and Duan, Haodong and Wang, Jiaqi and Qiao, Yu and Lin, Dahua and others},
  journal={arXiv:2403.20330},
  year={2024}
}

@inproceedings{gpt3,
  title={Language models are few-shot learners},
  author={Brown, Tom and Mann, Benjamin and Ryder, Nick and Subbiah, Melanie and Kaplan, Jared D and Dhariwal, Prafulla and Neelakantan, Arvind and Shyam, Pranav and Sastry, Girish and Askell, Amanda and others},
  booktitle={NeurIPS},
  year={2020}
}

@article{llava,
  title={Visual instruction tuning},
  author={Liu, Haotian and Li, Chunyuan and Wu, Qingyang and Lee, Yong Jae},
  journal={arXiv:2304.08485},
  year={2023}
}

@article{minigpt-4,
  title={Minigpt-4: Enhancing vision-language understanding with advanced large language models},
  author={Zhu, Deyao and Chen, Jun and Shen, Xiaoqian and Li, Xiang and Elhoseiny, Mohamed},
  journal={arXiv:2304.10592},
  year={2023}
}

@article{yue2023mmmu,
  title={Mmmu: A massive multi-discipline multimodal understanding and reasoning benchmark for expert agi},
  author={Yue, Xiang and Ni, Yuansheng and Zhang, Kai and Zheng, Tianyu and Liu, Ruoqi and Zhang, Ge and Stevens, Samuel and Jiang, Dongfu and Ren, Weiming and Sun, Yuxuan and others},
  journal={arXiv:2311.16502},
  year={2023}
}

@inproceedings{refcoco,
  title={Referitgame: Referring to objects in photographs of natural scenes},
  author={Kazemzadeh, Sahar and Ordonez, Vicente and Matten, Mark and Berg, Tamara},
  booktitle={EMNLP},
  year={2014}
}

@inproceedings{kembhavi2016diagram,
  title={A diagram is worth a dozen images},
  author={Kembhavi, Aniruddha and Salvato, Mike and Kolve, Eric and Seo, Minjoon and Hajishirzi, Hannaneh and Farhadi, Ali},
  booktitle={ECCV},
  year={2016},
}

@article{llama2,
  title={Llama 2: Open foundation and fine-tuned chat models},
  author={Touvron, Hugo and Martin, Louis and Stone, Kevin and Albert, Peter and Almahairi, Amjad and Babaei, Yasmine and Bashlykov, Nikolay and Batra, Soumya and Bhargava, Prajjwal and Bhosale, Shruti and others},
  journal={arXiv:2307.09288},
  year={2023}
}

@article{qwen2,
  title={Qwen2 technical report},
  author={Yang, An and Yang, Baosong and Hui, Binyuan and Zheng, Bo and Yu, Bowen and Zhou, Chang and Li, Chengpeng and Li, Chengyuan and Liu, Dayiheng and Huang, Fei and others},
  journal={arXiv:2407.10671},
  year={2024}
}

@inproceedings{mathvista,
  title={MathVista: Evaluating Mathematical Reasoning of Foundation Models in Visual Contexts},
  author={Pan Lu and Hritik Bansal and Tony Xia and Jiacheng Liu and Chunyuan Li and Hannaneh Hajishirzi and Hao Cheng and Kai{-}Wei Chang and Michel Galley and Jianfeng Gao},
  booktitle={ICLR},
  year={2024}
}

@article{mathvision,
  title={Measuring Multimodal Mathematical Reasoning with MATH-Vision Dataset}, 
  author={Ke Wang and Junting Pan and Weikang Shi and Zimu Lu and Mingjie Zhan and Hongsheng Li},
  journal={arXiv:2402.14804},
  year={2024}
}

@article{mme-realworld,
  title={MME-RealWorld: Could Your Multimodal LLM Challenge High-Resolution Real-World Scenarios that are Difficult for Humans?},
  author={Zhang, Yi-Fan and Zhang, Huanyu and Tian, Haochen and Fu, Chaoyou and Zhang, Shuangqing and Wu, Junfei and Li, Feng and Wang, Kun and Wen, Qingsong and Zhang, Zhang and others},
  journal={arXiv preprint arXiv:2408.13257},
  year={2024}
}

@article{mmmupro,
  title={MMMU-Pro: A More Robust Multi-discipline Multimodal Understanding Benchmark},
  author={Yue, Xiang and Zheng, Tianyu and Ni, Yuansheng and Wang, Yubo and Zhang, Kai and Tong, Shengbang and Sun, Yuxuan and Yin, Ming and Yu, Botao and Zhang, Ge and others},
  journal={arXiv preprint arXiv:2409.02813},
  year={2024}
}

@article{Liu_2024_OCRBench,
    title={OCRBench: on the hidden mystery of OCR in large multimodal models},
    volume={67},
    ISSN={1869-1919},
    url={http://dx.doi.org/10.1007/s11432-024-4235-6},
    DOI={10.1007/s11432-024-4235-6},
    number={12},
    journal={Science China Information Sciences},
    publisher={Springer Science and Business Media LLC},
    author={Liu, Yuliang and Li, Zhang and Huang, Mingxin and Yang, Biao and Yu, Wenwen and Li, Chunyuan and Yin, Xu-Cheng and Liu, Cheng-Lin and Jin, Lianwen and Bai, Xiang},
    year={2024},
    month=dec }

@article{wang2024charxiv,
  title={CharXiv: Charting Gaps in Realistic Chart Understanding in Multimodal LLMs},
  author={Wang, Zirui and Xia, Mengzhou and He, Luxi and Chen, Howard and Liu, Yitao and Zhu, Richard and Liang, Kaiqu and Wu, Xindi and Liu, Haotian and Malladi, Sadhika and Chevalier, Alexis and Arora, Sanjeev and Chen, Danqi},
  journal={arXiv preprint arXiv:2406.18521},
  year={2024}
}

@inproceedings{Librispeech,
  author       = {Vassil Panayotov and
                  Guoguo Chen and
                  Daniel Povey and
                  Sanjeev Khudanpur},
  title        = {Librispeech: An {ASR} corpus based on public domain audio books},
  booktitle    = {2015 {IEEE} International Conference on Acoustics, Speech and Signal
                  Processing, {ICASSP} 2015, South Brisbane, Queensland, Australia,
                  April 19-24, 2015},
  publisher    = {{IEEE}},
  year         = {2015},
}

@article{Conneau2022FLEURSFL,
  title={FLEURS: FEW-Shot Learning Evaluation of Universal Representations of Speech},
  author={Alexis Conneau and Min Ma and Simran Khanuja and Yu Zhang and Vera Axelrod and Siddharth Dalmia and Jason Riesa and Clara Rivera and Ankur Bapna},
  journal={2022 IEEE Spoken Language Technology Workshop (SLT)},
  year={2022},
  pages={798-805},
  url={https://api.semanticscholar.org/CorpusID:249062909}
}

@article{chen2024voicebench,
  title={Voicebench: Benchmarking llm-based voice assistants},
  author={Chen, Yiming and Yue, Xianghu and Zhang, Chen and Gao, Xiaoxue and Tan, Robby T and Li, Haizhou},
  journal={arXiv preprint arXiv:2410.17196},
  year={2024}
}

@article{qwen2-audio,
  title={Qwen2-audio technical report},
  author={Chu, Yunfei and Xu, Jin and Yang, Qian and Wei, Haojie and Wei, Xipin and Guo, Zhifang and Leng, Yichong and Lv, Yuanjun and He, Jinzheng and Lin, Junyang and others},
  journal={arXiv preprint arXiv:2407.10759},
  year={2024}
}

@article{seedtts,
  title={Seed-TTS: A Family of High-Quality Versatile Speech Generation Models},
  author={Anastassiou, Philip and Chen, Jiawei and Chen, Jitong and Chen, Yuanzhe and Chen, Zhuo and Chen, Ziyi and Cong, Jian and Deng, Lelai and Ding, Chuang and Gao, Lu and others},
  journal={arXiv preprint arXiv:2406.02430},
  year={2024}
}

@article{cosyvoice2,
  title={CosyVoice 2: Scalable Streaming Speech Synthesis with Large Language Models},
  author={Du, Zhihao and Wang, Yuxuan and Chen, Qian and Shi, Xian and Lv, Xiang and Zhao, Tianyu and Gao, Zhifu and Yang, Yexin and Gao, Changfeng and Wang, Hui and others},
  journal={arXiv preprint arXiv:2412.10117},
  year={2024}
}

@article{maskgct,
  title={Maskgct: Zero-shot text-to-speech with masked generative codec transformer},
  author={Wang, Yuancheng and Zhan, Haoyue and Liu, Liwei and Zeng, Ruihong and Guo, Haotian and Zheng, Jiachen and Zhang, Qiang and Zhang, Xueyao and Zhang, Shunsi and Wu, Zhizheng},
  journal={arXiv preprint arXiv:2409.00750},
  year={2024}
}

@article{f5tts,
  title={F5-tts: A fairytaler that fakes fluent and faithful speech with flow matching},
  author={Chen, Yushen and Niu, Zhikang and Ma, Ziyang and Deng, Keqi and Wang, Chunhui and Zhao, Jian and Yu, Kai and Chen, Xie},
  journal={arXiv preprint arXiv:2410.06885},
  year={2024}
}

@inproceedings{e2tts,
  title={E2 tts: Embarrassingly easy fully non-autoregressive zero-shot tts},
  author={Eskimez, Sefik Emre and Wang, Xiaofei and Thakker, Manthan and Li, Canrun and Tsai, Chung-Hsien and Xiao, Zhen and Yang, Hemin and Zhu, Zirun and Tang, Min and Tan, Xu and others},
  booktitle={2024 IEEE Spoken Language Technology Workshop (SLT)},
  pages={682--689},
  year={2024},
  organization={IEEE}
}

@article{sparktts,
  author       = {Xinsheng Wang and
                  Mingqi Jiang and
                  Ziyang Ma and
                  Ziyu Zhang and
                  Songxiang Liu and
                  Linqin Li and
                  Zheng Liang and
                  Qixi Zheng and
                  Rui Wang and
                  Xiaoqin Feng and
                  Weizhen Bian and
                  Zhen Ye and
                  Sitong Cheng and
                  Ruibin Yuan and
                  Zhixian Zhao and
                  Xinfa Zhu and
                  Jiahao Pan and
                  Liumeng Xue and
                  Pengcheng Zhu and
                  Yunlin Chen and
                  Zhifei Li and
                  Xie Chen and
                  Lei Xie and
                  Yike Guo and
                  Wei Xue},
  title        = {Spark-TTS: An Efficient LLM-Based Text-to-Speech Model with Single-Stream
                  Decoupled Speech Tokens},
  journal      = {CoRR},
  volume       = {abs/2503.01710},
  year         = {2025},
}

@article{cosyvoice3,
  author       = {Zhihao Du and
                  Changfeng Gao and
                  Yuxuan Wang and
                  Fan Yu and
                  Tianyu Zhao and
                  Hao Wang and
                  Xiang Lv and
                  Hui Wang and
                  Chongjia Ni and
                  Xian Shi and
                  Keyu An and
                  Guanrou Yang and
                  Yabin Li and
                  Yanni Chen and
                  Zhifu Gao and
                  Qian Chen and
                  Yue Gu and
                  Mengzhe Chen and
                  Yafeng Chen and
                  Shiliang Zhang and
                  Wen Wang and
                  Jieping Ye},
  title        = {CosyVoice 3: Towards In-the-wild Speech Generation via Scaling-up
                  and Post-training},
  journal      = {CoRR},
  volume       = {abs/2505.17589},
  year         = {2025}
}

@article{mimoaudio,
  title={MiMo-Audio: Audio Language Models are Few-Shot Learners},
  author={Xiaomi LLM-Core Team Dong Zhang and Gang Wang and Jinlong Xue and Kai Fang and Liang Zhao and Rui Ma and Shu-Qin Ren and Shuo Liu and Tao Guo and Weiji Zhuang and Xin Zhang and Xi-Na Song and Yihan Yan and Yongzhe He and Cici and Bowen Shen and Chengxuan Zhu and Chong Ma and Chun Chen and Heyu Chen and Jiawei Li and Lei Li and Menghang Zhu and Peidian Li and Qiying Wang and Sirui Deng and Weimin Xiong and Wen Huang and Wenyu Yang and Yilin Jiang and Yixin Yang and Yu-Shi Tian and Yue Ma and Yue Yu and Zihan Zhang and Zihao Yue and Bangjun Xiao and Bin Xia and Bofei Gao and Bowen Ye and Can Cai and Chang Liu and Chenhong He and Chunan Li and Dawei Zhu and Duo Zhang and Fengyuan Shi and Guoan Wang and Hailin Zhang and Hanglong Lv and Hanyu Li and Hao Tian and Hengxu Qu and Hong-Mei Xu and Houbin Zhang and Huaqiu Liu and Jiangshan Duo and Jia Zuo and Jianyu Wei and Jiebao Xiao and Jinhao Dong and Jun Shi and Junhao Hu and Kainan Bao and Kang Zhou and Linghao Zhang and Meng Chen and Nuo Chen and Peng Zhang and Qian Chen and Qiantong Wang and Rang Li and Shao-yang Liu and Shengfan Wang and Shicheng Li and Shi-liang Yu and Shijie Cao and Shimao Chen and Shuhao Gu and Weikun Wang and Wen-Juan Ma and Xia Deng and Xing Yong and Xing Zhang and Xu Wang and Yi-Hao Song and Yihao Zhao and Yingbo Zhao and Yizhao Gao and Yu Cheng and Yuanfang Tu and Yudong Wang and Zhaojun Huang and Zheng-Yu Tang and Zhenrui Lin and Zhichao Song and Zhi-Yue Xu and Zhixian Zheng and Zi-Cheng Jiang},
  journal={ArXiv},
  year={2025},
  volume={abs/2512.23808},
  url={https://api.semanticscholar.org/CorpusID:284351195}
}

@article{minimaxspeech,
  author       = {Bowen Zhang and
                  Congchao Guo and
                  Geng Yang and
                  Hang Yu and
                  Haozhe Zhang and
                  Heidi Lei and
                  Jialong Mai and
                  Junjie Yan and
                  Kaiyue Yang and
                  Mingqi Yang and
                  Peikai Huang and
                  Ruiyang Jin and
                  Sitan Jiang and
                  Weihua Cheng and
                  Yawei Li and
                  Yichen Xiao and
                  Yiying Zhou and
                  Yongmao Zhang and
                  Yuan Lu and
                  Yucen He},
  title        = {MiniMax-Speech: Intrinsic Zero-Shot Text-to-Speech with a Learnable
                  Speaker Encoder},
  journal      = {CoRR},
  volume       = {abs/2505.07916},
  year         = {2025}
}

@misc{sakshi2024mmaumassivemultitaskaudio,
      title={MMAU: A Massive Multi-Task Audio Understanding and Reasoning Benchmark}, 
      author={S Sakshi and Utkarsh Tyagi and Sonal Kumar and Ashish Seth and Ramaneswaran Selvakumar and Oriol Nieto and Ramani Duraiswami and Sreyan Ghosh and Dinesh Manocha},
      year={2024},
      eprint={2410.19168},
      archivePrefix={arXiv},
      primaryClass={eess.AS},
      url={https://arxiv.org/abs/2410.19168}, 
}

@article{qwen2.5omni,
  title={Qwen2. 5-omni technical report},
  author={Xu, Jin and Guo, Zhifang and He, Jinzheng and Hu, Hangrui and He, Ting and Bai, Shuai and Chen, Keqin and Wang, Jialin and Fan, Yang and Dang, Kai and others},
  journal={arXiv preprint arXiv:2503.20215},
  year={2025}
}

@article{qwen2.5vl,
  title={Qwen2. 5-vl technical report},
  author={Bai, Shuai and Chen, Keqin and Liu, Xuejing and Wang, Jialin and Ge, Wenbin and Song, Sibo and Dang, Kai and Wang, Peng and Wang, Shijie and Tang, Jun and others},
  journal={arXiv preprint arXiv:2502.13923},
  year={2025}
}

@article{qwen3,
  title={Qwen3 technical report},
  author={Yang, An and Li, Anfeng and Yang, Baosong and Zhang, Beichen and Hui, Binyuan and Zheng, Bo and Yu, Bowen and Gao, Chang and Huang, Chengen and Lv, Chenxu and others},
  journal={arXiv preprint arXiv:2505.09388},
  year={2025}
}

@article{gemini2.5,
  title={Gemini 2.5: Pushing the frontier with advanced reasoning, multimodality, long context, and next generation agentic capabilities},
  author={Comanici, Gheorghe and Bieber, Eric and Schaekermann, Mike and Pasupat, Ice and Sachdeva, Noveen and Dhillon, Inderjit and Blistein, Marcel and Ram, Ori and Zhang, Dan and Rosen, Evan and others},
  journal={arXiv preprint arXiv:2507.06261},
  year={2025}
}

@misc{bfcl,
    title={Berkeley Function Calling Leaderboard}, 
    author={Fanjia Yan and Huanzhi Mao and Charlie Cheng-Jie Ji
    and Tianjun Zhang and Shishir G. Patil and Ion Stoica and Joseph E.
    Gonzalez},
    howpublished={\url{https://gorilla.cs.berkeley.edu/blogs/8_berkeley_function_calling_leaderboard.html}},
    year={2024},
}

@article{gspo,
  title={Group sequence policy optimization},
  author={Zheng, Chujie and Liu, Shixuan and Li, Mingze and Chen, Xiong-Hui and Yu, Bowen and Gao, Chang and Dang, Kai and Liu, Yuqiong and Men, Rui and Yang, An and others},
  journal={arXiv preprint arXiv:2507.18071},
  year={2025}
}

@article{zang2025you,
  title={Are you really listening? boosting perceptual awareness in music-qa benchmarks},
  author={Zang, Yongyi and O'Brien, Sean and Berg-Kirkpatrick, Taylor and McAuley, Julian and Novack, Zachary},
  journal={arXiv preprint arXiv:2504.00369},
  year={2025}
}

@article{mmar,
  author       = {Ziyang Ma and
                  Yinghao Ma and
                  Yanqiao Zhu and
                  Chen Yang and
                  Yi{-}Wen Chao and
                  Ruiyang Xu and
                  Wenxi Chen and
                  Yuanzhe Chen and
                  Zhuo Chen and
                  Jian Cong and
                  Kai Li and
                  Keliang Li and
                  Siyou Li and
                  Xinfeng Li and
                  Xiquan Li and
                  Zheng Lian and
                  Yuzhe Liang and
                  Minghao Liu and
                  Zhikang Niu and
                  Tianrui Wang and
                  Yuping Wang and
                  Yuxuan Wang and
                  Yihao Wu and
                  Guanrou Yang and
                  Jianwei Yu and
                  Ruibin Yuan and
                  Zhisheng Zheng and
                  Ziya Zhou and
                  Haina Zhu and
                  Wei Xue and
                  Emmanouil Benetos and
                  Kai Yu and
                  Chng Eng Siong and
                  Xie Chen},
  title        = {{MMAR:} {A} Challenging Benchmark for Deep Reasoning in Speech, Audio,
                  Music, and Their Mix},
  journal      = {CoRR},
  volume       = {abs/2505.13032},
  year         = {2025},
  url          = {https://doi.org/10.48550/arXiv.2505.13032},
  doi          = {10.48550/ARXIV.2505.13032},
  eprinttype    = {arXiv},
  eprint       = {2505.13032},
  bibsource    = {dblp computer science bibliography, https://dblp.org}
}

@article{mmsu,
  author       = {Dingdong Wang and
                  Jincenzi Wu and
                  Junan Li and
                  Dongchao Yang and
                  Xueyuan Chen and
                  Tianhua Zhang and
                  Helen Meng},
  title        = {{MMSU:} {A} Massive Multi-task Spoken Language Understanding and Reasoning
                  Benchmark},
  journal      = {CoRR},
  volume       = {abs/2506.04779},
  year         = {2025},
  url          = {https://doi.org/10.48550/arXiv.2506.04779},
  doi          = {10.48550/ARXIV.2506.04779},
  eprinttype    = {arXiv},
  eprint       = {2506.04779},
  bibsource    = {dblp computer science bibliography, https://dblp.org}
}

@article{supergpqa,
  author       = {M.{-}A{-}P. Team and
                  Xinrun Du and
                  Yifan Yao and
                  Kaijing Ma and
                  Bingli Wang and
                  Tianyu Zheng and
                  Kang Zhu and
                  Minghao Liu and
                  Yiming Liang and
                  Xiaolong Jin and
                  Zhenlin Wei and
                  Chujie Zheng and
                  Kaixin Deng and
                  Shian Jia and
                  Sichao Jiang and
                  Yiyan Liao and
                  Rui Li and
                  Qinrui Li and
                  Sirun Li and
                  Yizhi Li and
                  Yunwen Li and
                  Dehua Ma and
                  Yuansheng Ni and
                  Haoran Que and
                  Qiyao Wang and
                  Zhoufutu Wen and
                  Siwei Wu and
                  Tianshun Xing and
                  Ming Xu and
                  Zhenzhu Yang and
                  Zekun Moore Wang and
                  Jun Zhou and
                  Yuelin Bai and
                  Xingyuan Bu and
                  Chenglin Cai and
                  Liang Chen and
                  Yifan Chen and
                  Chengtuo Cheng and
                  Tianhao Cheng and
                  Keyi Ding and
                  Siming Huang and
                  Yun Huang and
                  Yaoru Li and
                  Yizhe Li and
                  Zhaoqun Li and
                  Tianhao Liang and
                  Chengdong Lin and
                  Hongquan Lin and
                  Yinghao Ma and
                  Tianyang Pang and
                  Zhongyuan Peng and
                  Zifan Peng and
                  Qige Qi and
                  Shi Qiu and
                  Xingwei Qu and
                  Shanghaoran Quan and
                  Yizhou Tan and
                  Zili Wang and
                  Chenqing Wang and
                  Hao Wang and
                  Yiya Wang and
                  Yubo Wang and
                  Jiajun Xu and
                  Kexin Yang and
                  Ruibin Yuan and
                  Yuanhao Yue and
                  Tianyang Zhan and
                  Chun Zhang and
                  Jinyang Zhang and
                  Xiyue Zhang and
                  Xingjian Zhang and
                  Yue Zhang and
                  Yongchi Zhao and
                  Xiangyu Zheng and
                  Chenghua Zhong and
                  Yang Gao and
                  Zhoujun Li and
                  Dayiheng Liu and
                  Qian Liu and
                  Tianyu Liu and
                  Shiwen Ni and
                  Junran Peng and
                  Yujia Qin and
                  Wenbo Su and
                  Guoyin Wang and
                  Shi Wang and
                  Jian Yang and
                  Min Yang and
                  Meng Cao and
                  Xiang Yue and
                  Zhaoxiang Zhang and
                  Wangchunshu Zhou and
                  Jiaheng Liu and
                  Qunshu Lin and
                  Wenhao Huang and
                  Ge Zhang},
  title        = {SuperGPQA: Scaling {LLM} Evaluation across 285 Graduate Disciplines},
  journal      = {CoRR},
  volume       = {abs/2502.14739},
  year         = {2025},
  url          = {https://doi.org/10.48550/arXiv.2502.14739},
  doi          = {10.48550/ARXIV.2502.14739},
  eprinttype    = {arXiv},
  eprint       = {2502.14739},
  bibsource    = {dblp computer science bibliography, https://dblp.org}
}

@misc{barres2025tau2,
      title={$\tau^2$-Bench: Evaluating Conversational Agents in a Dual-Control Environment}, 
      author={Victor Barres and Honghua Dong and Soham Ray and Xujie Si and Karthik Narasimhan},
      year={2025},
      eprint={2506.07982},
      archivePrefix={arXiv},
      primaryClass={cs.AI},
      url={https://arxiv.org/abs/2506.07982}, 
}

@inproceedings{countbench,
  author       = {Roni Paiss and
                  Ariel Ephrat and
                  Omer Tov and
                  Shiran Zada and
                  Inbar Mosseri and
                  Michal Irani and
                  Tali Dekel},
  title        = {Teaching {CLIP} to Count to Ten},
  booktitle    = {{IEEE/CVF} International Conference on Computer Vision, {ICCV} 2023,
                  Paris, France, October 1-6, 2023},
  pages        = {3147--3157},
  publisher    = {{IEEE}},
  year         = {2023},
}

@inproceedings{hallusionbench,
  author       = {Tianrui Guan and
                  Fuxiao Liu and
                  Xiyang Wu and
                  Ruiqi Xian and
                  Zongxia Li and
                  Xiaoyu Liu and
                  Xijun Wang and
                  Lichang Chen and
                  Furong Huang and
                  Yaser Yacoob and
                  Dinesh Manocha and
                  Tianyi Zhou},
  title        = {Hallusionbench: An Advanced Diagnostic Suite for Entangled Language
                  Hallucination and Visual Illusion in Large Vision-Language Models},
  booktitle    = {{IEEE/CVF} Conference on Computer Vision and Pattern Recognition,
                  {CVPR} 2024, Seattle, WA, USA, June 16-22, 2024},
  pages        = {14375--14385},
  publisher    = {{IEEE}},
  year         = {2024},
}

@article{lvbench,
  author       = {Weihan Wang and
                  Zehai He and
                  Wenyi Hong and
                  Yean Cheng and
                  Xiaohan Zhang and
                  Ji Qi and
                  Shiyu Huang and
                  Bin Xu and
                  Yuxiao Dong and
                  Ming Ding and
                  Jie Tang},
  title        = {LVBench: An Extreme Long Video Understanding Benchmark},
  journal      = {CoRR},
  volume       = {abs/2406.08035},
  year         = {2024},
}

@inproceedings{mlvu,
  author       = {Junjie Zhou and
                  Yan Shu and
                  Bo Zhao and
                  Boya Wu and
                  Zhengyang Liang and
                  Shitao Xiao and
                  Minghao Qin and
                  Xi Yang and
                  Yongping Xiong and
                  Bo Zhang and
                  Tiejun Huang and
                  Zheng Liu},
  title        = {{MLVU:} Benchmarking Multi-task Long Video Understanding},
  booktitle    = {{IEEE/CVF} Conference on Computer Vision and Pattern Recognition,
                  {CVPR} 2025, Nashville, TN, USA, June 11-15, 2025},
  pages        = {13691--13701},
  publisher    = {Computer Vision Foundation / {IEEE}},
  year         = {2025},
}

@article{dailyomni,
  author       = {Ziwei Zhou and
                  Rui Wang and
                  Zuxuan Wu},
  title        = {Daily-Omni: Towards Audio-Visual Reasoning with Temporal Alignment
                  across Modalities},
  journal      = {CoRR},
  volume       = {abs/2505.17862},
  year         = {2025},
}

@article{worldsense,
  author       = {Jack Hong and
                  Shilin Yan and
                  Jiayin Cai and
                  Xiaolong Jiang and
                  Yao Hu and
                  Weidi Xie},
  title        = {WorldSense: Evaluating Real-world Omnimodal Understanding for Multimodal
                  LLMs},
  journal      = {CoRR},
  volume       = {abs/2502.04326},
  year         = {2025},
}

@article{omnigaia,
  title={OmniGAIA: Towards Native Omni-Modal AI Agents},
  author={Li, Xiaoxi and Jiao, Wenxiang and Jin, Jiarui and Wang, Shijian and Dong, Guanting and Jin, Jiajie and Wang, Hao and Wang, Yinuo and Wen, Ji-Rong and Lu, Yuan and others},
  journal={arXiv preprint arXiv:2602.22897},
  year={2026}
}

@article{avspeakerbench,
  title={See, Hear, and Understand: Benchmarking Audiovisual Human Speech Understanding in Multimodal Large Language Models},
  author={Nguyen, Le Thien Phuc and Yu, Zhuoran and Hang, Samuel Low Yu and An, Subin and Lee, Jeongik and Ban, Yohan and Chung, SeungEun and Nguyen, Thanh-Huy and Maeng, JuWan and Lee, Soochahn and others},
  journal={arXiv preprint arXiv:2512.02231},
  year={2025}
}

@inproceedings{avut,
  title={Audio-centric video understanding benchmark without text shortcut},
  author={Yang, Yudong and Zhuang, Jimin and Sun, Guangzhi and Tang, Changli and Li, Yixuan and Li, Peihan and Jiang, Yifan and Li, Wei and Ma, Zejun and Zhang, Chao},
  booktitle={Proceedings of the 2025 Conference on Empirical Methods in Natural Language Processing},
  pages={6580--6598},
  year={2025}
}

@article{qivd,
  title={Can Vision-Language Models Answer Face to Face Questions in the Real-World?},
  author={Pourreza, Reza and Dagli, Rishit and Bhattacharyya, Apratim and Panchal, Sunny and Berger, Guillaume and Memisevic, Roland},
  journal={arXiv preprint arXiv:2503.19356},
  year={2025}
}

@inproceedings{videomme,
  title={Video-mme: The first-ever comprehensive evaluation benchmark of multi-modal llms in video analysis},
  author={Fu, Chaoyou and Dai, Yuhan and Luo, Yongdong and Li, Lei and Ren, Shuhuai and Zhang, Renrui and Wang, Zihan and Zhou, Chenyu and Shen, Yunhang and Zhang, Mengdan and others},
  booktitle={Proceedings of the IEEE/CVF conference on computer vision and pattern recognition},
  pages={24108--24118},
  year={2025}
}

@article{omnicloze,
  title={Omni-Captioner: Data Pipeline, Models, and Benchmark for Omni Detailed Perception},
  author={Ma, Ziyang and Xu, Ruiyang and Xing, Zhenghao and Chu, Yunfei and Wang, Yuxuan and He, Jinzheng and Xu, Jin and Heng, Pheng-Ann and Yu, Kai and Lin, Junyang and others},
  journal={arXiv preprint arXiv:2510.12720},
  year={2025}
}

@article{ifbench,
  author       = {Valentina Pyatkin and
                  Saumya Malik and
                  Victoria Graf and
                  Hamish Ivison and
                  Shengyi Huang and
                  Pradeep Dasigi and
                  Nathan Lambert and
                  Hannaneh Hajishirzi},
  title        = {Generalizing Verifiable Instruction Following},
  journal      = {CoRR},
  volume       = {abs/2507.02833},
  year         = {2025},
  url          = {https://doi.org/10.48550/arXiv.2507.02833},
  doi          = {10.48550/ARXIV.2507.02833},
  eprinttype   = {arXiv},
  eprint       = {2507.02833},
  bibsource    = {dblp computer science bibliography, https://dblp.org}
}

@misc{aalcr,
  title={Artificial Analysis Long Context Reasoning Benchmark (LCR)},
  author={Artificial Analysis Team},
  year={2025},
  howpublished={Artificial Analysis, Inc.},
  note={Dataset}
}

@inproceedings{longbenchv2,
  author       = {Yushi Bai and
                  Shangqing Tu and
                  Jiajie Zhang and
                  Hao Peng and
                  Xiaozhi Wang and
                  Xin Lv and
                  Shulin Cao and
                  Jiazheng Xu and
                  Lei Hou and
                  Yuxiao Dong and
                  Jie Tang and
                  Juanzi Li},
  editor       = {Wanxiang Che and
                  Joyce Nabende and
                  Ekaterina Shutova and
                  Mohammad Taher Pilehvar},
  title        = {LongBench v2: Towards Deeper Understanding and Reasoning on Realistic
                  Long-context Multitasks},
  booktitle    = {Proceedings of the 63rd Annual Meeting of the Association for Computational
                  Linguistics (Volume 1: Long Papers), {ACL} 2025, Vienna, Austria,
                  July 27 - August 1, 2025},
  pages        = {3639--3664},
  publisher    = {Association for Computational Linguistics},
  year         = {2025},
  url          = {https://aclanthology.org/2025.acl-long.183/},
  bibsource    = {dblp computer science bibliography, https://dblp.org}
}

@misc{hmmtnov25,
  title = {MathArena: Evaluating LLMs on Uncontaminated Math Competitions},
  author = {Mislav Balunović and Jasper Dekoninck and Ivo Petrov and Nikola Jovanović and Martin Vechev},
  copyright = {MIT},
  url = {https://matharena.ai/},
  publisher = {SRI Lab, ETH Zurich},
  month = feb,
  year = {2025},
}

@inproceedings{imoanswerbench,
    title = "Towards Robust Mathematical Reasoning",
    author  = {Thang Luong and Dawsen Hwang and Hoang H. Nguyen and Golnaz Ghiasi and Yuri Chervonyi and Insuk Seo and Junsu Kim and Garrett Bingham and Jonathan Lee and Swaroop Mishra and Alex Zhai and Clara Huiyi Hu and Henryk Michalewski and Jimin Kim and Jeonghyun Ahn and Junhwi Bae and Xingyou Song and Trieu H. Trinh and Quoc V. Le and Junehyuk Jung},
    booktitle = "Proceedings of the 2025 Conference on Empirical Methods in Natural Language Processing",
    year = "2025",
    url = "https://aclanthology.org/2025.emnlp-main.1794/",
}

@misc{hao2025songformer,
  title         = {SongFormer: Scaling Music Structure Analysis with Heterogeneous Supervision},
  author        = {Chunbo Hao and Ruibin Yuan and Jixun Yao and Qixin Deng and Xinyi Bai and Wei Xue and Lei Xie},
  year          = {2025},
  eprint        = {2510.02797},
  archivePrefix = {arXiv},
  primaryClass  = {eess.AS},
  url           = {https://arxiv.org/abs/2510.02797}
}

@misc{zhang2025wildspeechbenchbenchmarkingendtoendspeechllms,
     title={WildSpeech-Bench: Benchmarking End-to-End SpeechLLMs in the Wild}, 
     author={Linhao Zhang and Jian Zhang and Bokai Lei and Chuhan Wu and Aiwei Liu and Wei Jia and Xiao Zhou},
     year={2025},
     eprint={2506.21875},
     archivePrefix={arXiv},
     primaryClass={cs.CL},
}

@article{jiang2025speechrole,
  title={SpeechRole: A Large-Scale Dataset and Benchmark for Evaluating Speech Role-Playing Agents},
  author={Jiang, Changhao and Sun, Jiajun and Cao, Yifei and Zhuang, Jiabao and Li, Hui and Fan, Xiaoran and Zhang, Ming and Ye, Junjie and Dou, Shihan and Xi, Zhiheng and others},
  journal={arXiv preprint arXiv:2508.02013},
  year={2025}
}

@article{yan2025uro,
  title={URO-Bench: A Comprehensive Benchmark for End-to-End Spoken Dialogue Models},
  author={Yan, Ruiqi and Li, Xiquan and Chen, Wenxi and Niu, Zhikang and Yang, Chen and Ma, Ziyang and Yu, Kai and Chen, Xie},
  journal={arXiv preprint arXiv:2502.17810},
  year={2025}
}

@inproceedings{DBLP:conf/icassp/ZhangLGSYXXBCZW22,
  author       = {Binbin Zhang and
                  Hang Lv and
                  Pengcheng Guo and
                  Qijie Shao and
                  Chao Yang and
                  Lei Xie and
                  Xin Xu and
                  Hui Bu and
                  Xiaoyu Chen and
                  Chenchen Zeng and
                  Di Wu and
                  Zhendong Peng},
  title        = {{WENETSPEECH:} {A} 10000+ Hours Multi-Domain Mandarin Corpus for Speech
                  Recognition},
  booktitle    = {{IEEE} International Conference on Acoustics, Speech and Signal Processing,
                  {ICASSP} 2022, Virtual and Singapore, 23-27 May 2022},
  pages        = {6182--6186},
  publisher    = {{IEEE}},
  year         = {2022},
  url          = {https://doi.org/10.1109/ICASSP43922.2022.9746682},
  doi          = {10.1109/ICASSP43922.2022.9746682},
  bibsource    = {dblp computer science bibliography, https://dblp.org}
}

@inproceedings{DBLP:conf/interspeech/WangWZWLXZXB22,
  author       = {Yu Wang and
                  Xinsheng Wang and
                  Pengcheng Zhu and
                  Jie Wu and
                  Hanzhao Li and
                  Heyang Xue and
                  Yongmao Zhang and
                  Lei Xie and
                  Mengxiao Bi},
  editor       = {Hanseok Ko and
                  John H. L. Hansen},
  title        = {Opencpop: {A} High-Quality Open Source Chinese Popular Song Corpus
                  for Singing Voice Synthesis},
  booktitle    = {23rd Annual Conference of the International Speech Communication Association,
                  Interspeech 2022, Incheon, Korea, September 18-22, 2022},
  pages        = {4242--4246},
  publisher    = {{ISCA}},
  year         = {2022},
  url          = {https://doi.org/10.21437/Interspeech.2022-48},
  doi          = {10.21437/INTERSPEECH.2022-48},
  bibsource    = {dblp computer science bibliography, https://dblp.org}
}

@inproceedings{DBLP:conf/nips/Tang0XSLZWTXZYL21,
  author       = {Zhiyuan Tang and
                  Dong Wang and
                  Yanguang Xu and
                  Jianwei Sun and
                  Xiaoning Lei and
                  Shuaijiang Zhao and
                  Cheng Wen and
                  Xingjun Tan and
                  Chuandong Xie and
                  Shuran Zhou and
                  Rui Yan and
                  Chenjia Lv and
                  Yang Han and
                  Wei Zou and
                  Xiangang Li},
  editor       = {Joaquin Vanschoren and
                  Sai{-}Kit Yeung},
  title        = {KeSpeech: An Open Source Speech Dataset of Mandarin and Its Eight
                  Subdialects},
  booktitle    = {Proceedings of the Neural Information Processing Systems Track on
                  Datasets and Benchmarks 1, NeurIPS Datasets and Benchmarks 2021, December
                  2021, virtual},
  year         = {2021},
  url          = {https://datasets-benchmarks-proceedings.neurips.cc/paper/2021/hash/0336dcbab05b9d5ad24f4333c7658a0e-Abstract-round2.html},
  bibsource    = {dblp computer science bibliography, https://dblp.org}
}

@article{DBLP:journals/taslp/HsuJ10,
  author       = {Chao{-}Ling Hsu and
                  Jyh{-}Shing Roger Jang},
  title        = {On the Improvement of Singing Voice Separation for Monaural Recordings
                  Using the {MIR-1K} Dataset},
  journal      = {{IEEE} Trans. Speech Audio Process.},
  volume       = {18},
  number       = {2},
  pages        = {310--319},
  year         = {2010},
  url          = {https://doi.org/10.1109/TASL.2009.2026503},
  doi          = {10.1109/TASL.2009.2026503},
  bibsource    = {dblp computer science bibliography, https://dblp.org}
}

@inproceedings{DBLP:conf/lrec/ArdilaBDKMHMSTW20,
  author       = {Rosana Ardila and
                  Megan Branson and
                  Kelly Davis and
                  Michael Kohler and
                  Josh Meyer and
                  Michael Henretty and
                  Reuben Morais and
                  Lindsay Saunders and
                  Francis M. Tyers and
                  Gregor Weber},
  editor       = {Nicoletta Calzolari and
                  Fr{\'{e}}d{\'{e}}ric B{\'{e}}chet and
                  Philippe Blache and
                  Khalid Choukri and
                  Christopher Cieri and
                  Thierry Declerck and
                  Sara Goggi and
                  Hitoshi Isahara and
                  Bente Maegaard and
                  Joseph Mariani and
                  H{\'{e}}l{\`{e}}ne Mazo and
                  Asunci{\'{o}}n Moreno and
                  Jan Odijk and
                  Stelios Piperidis},
  title        = {Common Voice: {A} Massively-Multilingual Speech Corpus},
  booktitle    = {Proceedings of The 12th Language Resources and Evaluation Conference,
                  {LREC} 2020, Marseille, France, May 11-16, 2020},
  pages        = {4218--4222},
  publisher    = {European Language Resources Association},
  year         = {2020},
  url          = {https://aclanthology.org/2020.lrec-1.520/},
  bibsource    = {dblp computer science bibliography, https://dblp.org}
}

@article{qwen3omni,
  title={Qwen3-Omni Technical Report},
  author={Jin Xu and Zhifang Guo and Hangrui Hu and Yunfei Chu and Xiong Wang and Jinzheng He and Yuxuan Wang and Xianzhong Shi and Ting He and Xinfa Zhu and Yuanjun Lv and Yongqi Wang and Dake Guo and He Wang and Linhan Ma and Pei Zhang and Xinyu Zhang and Hongkun Hao and Zishan Guo and Baosong Yang and Bin Zhang and Ziyang Ma and Xipin Wei and Shuai Bai and Ke Chen and Xue Lian Liu and Peng Wang and Ming Yang and Dayiheng Liu and Xingzhang Ren and Bo Zheng and Rui Men and Fan Zhou and Bowen Yu and Jianxin Yang and Le Yu and Jing-Jun Zhou and Junyang Lin},
  journal={ArXiv},
  year={2025},
  volume={abs/2509.17765},
}

@misc{qwen35blog,
    title = {Qwen3.5: Accelerating Productivity with Native Multimodal Agents},
    url = {https://qwen.ai/blog?id=qwen3.5},
    author = {Qwen Team},
    month = {February},
    year = {2026}
}

@inproceedings{dynamath,
  author       = {Chengke Zou and
                  Xingang Guo and
                  Rui Yang and
                  Junyu Zhang and
                  Bin Hu and
                  Huan Zhang},
  title        = {DynaMath: {A} Dynamic Visual Benchmark for Evaluating Mathematical
                  Reasoning Robustness of Vision Language Models},
  booktitle    = {ICLR},
  publisher    = {OpenReview.net},
  year         = {2025},
}

@article{zerobench,
  author       = {Jonathan Roberts and
                  Mohammad Reza Taesiri and
                  Ansh Sharma and
                  Akash Gupta and
                  Samuel Roberts and
                  Ioana Croitoru and
                  Simion{-}Vlad Bogolin and
                  Jialu Tang and
                  Florian Langer and
                  Vyas Raina and
                  Vatsal Raina and
                  Hanyi Xiong and
                  Vishaal Udandarao and
                  Jingyi Lu and
                  Shiyang Chen and
                  Sam Purkis and
                  Tianshuo Yan and
                  Wenye Lin and
                  Gyungin Shin and
                  Qiaochu Yang and
                  Anh Totti Nguyen and
                  Kai Han and
                  Samuel Albanie},
  title        = {ZeroBench: An Impossible Visual Benchmark for Contemporary Large Multimodal
                  Models},
  journal      = {CoRR},
  volume       = {abs/2502.09696},
  year         = {2025},
}

@article{simplevqa,
  author       = {Xianfu Cheng and
                  Wei Zhang and
                  Shiwei Zhang and
                  Jian Yang and
                  Xiangyuan Guan and
                  Xianjie Wu and
                  Xiang Li and
                  Ge Zhang and
                  Jiaheng Liu and
                  Yuying Mai and
                  Yutao Zeng and
                  Zhoufutu Wen and
                  Ke Jin and
                  Baorui Wang and
                  Weixiao Zhou and
                  Yunhong Lu and
                  Tongliang Li and
                  Wenhao Huang and
                  Zhoujun Li},
  title        = {SimpleVQA: Multimodal Factuality Evaluation for Multimodal Large Language
                  Models},
  journal      = {CoRR},
  volume       = {abs/2502.13059},
  year         = {2025}
}

@article{ccocr,
  author       = {Zhibo Yang and
                  Jun Tang and
                  Zhaohai Li and
                  Pengfei Wang and
                  Jianqiang Wan and
                  Humen Zhong and
                  Xuejing Liu and
                  Mingkun Yang and
                  Peng Wang and
                  Shuai Bai and
                  LianWen Jin and
                  Junyang Lin},
  title        = {{CC-OCR:} {A} Comprehensive and Challenging {OCR} Benchmark for Evaluating
                  Large Multimodal Models in Literacy},
  journal      = {CoRR},
  volume       = {abs/2412.02210},
  year         = {2024},
}

@inproceedings{mmlongbench,
  author       = {Yubo Ma and
                  Yuhang Zang and
                  Liangyu Chen and
                  Meiqi Chen and
                  Yizhu Jiao and
                  Xinze Li and
                  Xinyuan Lu and
                  Ziyu Liu and
                  Yan Ma and
                  Xiaoyi Dong and
                  Pan Zhang and
                  Liangming Pan and
                  Yu{-}Gang Jiang and
                  Jiaqi Wang and
                  Yixin Cao and
                  Aixin Sun},
  editor       = {Amir Globersons and
                  Lester Mackey and
                  Danielle Belgrave and
                  Angela Fan and
                  Ulrich Paquet and
                  Jakub M. Tomczak and
                  Cheng Zhang},
  title        = {{MMLONGBENCH-DOC:} Benchmarking Long-context Document Understanding
                  with Visualizations},
  booktitle    = {Advances in Neural Information Processing Systems 38: Annual Conference
                  on Neural Information Processing Systems 2024, NeurIPS 2024, Vancouver,
                  BC, Canada, December 10 - 15, 2024},
  year         = {2024}
}

@article{erqa,
  author       = {Gemini Robotics Team},
  title        = {Gemini Robotics: Bringing {AI} into the Physical World},
  journal      = {CoRR},
  volume       = {abs/2503.20020},
  year         = {2025},
}

@inproceedings{odinw,
  author       = {Liunian Harold Li and
                  Pengchuan Zhang and
                  Haotian Zhang and
                  Jianwei Yang and
                  Chunyuan Li and
                  Yiwu Zhong and
                  Lijuan Wang and
                  Lu Yuan and
                  Lei Zhang and
                  Jenq{-}Neng Hwang and
                  Kai{-}Wei Chang and
                  Jianfeng Gao},
  title        = {Grounded Language-Image Pre-training},
  booktitle    = {CVPR},
  pages        = {10955--10965},
  publisher    = {{IEEE}},
  year         = {2022}
}

@inproceedings{embspatial,
  author       = {Mengfei Du and
                  Binhao Wu and
                  Zejun Li and
                  Xuanjing Huang and
                  Zhongyu Wei},
  editor       = {Lun{-}Wei Ku and
                  Andre Martins and
                  Vivek Srikumar},
  title        = {EmbSpatial-Bench: Benchmarking Spatial Understanding for Embodied
                  Tasks with Large Vision-Language Models},
  booktitle    = {Proceedings of the 62nd Annual Meeting of the Association for Computational
                  Linguistics (Volume 2: Short Papers), {ACL} 2024, Bangkok, Thailand,
                  August 11-16, 2024},
  pages        = {346--355},
  publisher    = {Association for Computational Linguistics},
  year         = {2024},
}

@inproceedings{mmvu,
  author       = {Yilun Zhao and
                  Haowei Zhang and
                  Lujing Xie and
                  Tongyan Hu and
                  Guo Gan and
                  Yitao Long and
                  Zhiyuan Hu and
                  Weiyuan Chen and
                  Chuhan Li and
                  Zhijian Xu and
                  Chengye Wang and
                  Ziyao Shangguan and
                  Zhenwen Liang and
                  Yixin Liu and
                  Chen Zhao and
                  Arman Cohan},
  title        = {{MMVU:} Measuring Expert-Level Multi-Discipline Video Understanding},
  booktitle    = {{IEEE/CVF} Conference on Computer Vision and Pattern Recognition,
                  {CVPR} 2025, Nashville, TN, USA, June 11-15, 2025},
  pages        = {8475--8489},
  publisher    = {Computer Vision Foundation / {IEEE}},
  year         = {2025},
}

@article{videoocr,
  author       = {Yang Shi and
                  Huanqian Wang and
                  Wulin Xie and
                  Huanyao Zhang and
                  Lijie Zhao and
                  Yifan Zhang and
                  Xinfeng Li and
                  Chaoyou Fu and
                  Zhuoer Wen and
                  Wenting Liu and
                  Zhuoran Zhang and
                  Xinlong Chen and
                  Bohan Zeng and
                  Sihan Yang and
                  Yuanxing Zhang and
                  Pengfei Wan and
                  Haotian Wang and
                  Wenjing Yang},
  title        = {MME-VideoOCR: Evaluating OCR-Based Capabilities of Multimodal LLMs
                  in Video Scenarios},
  journal      = {CoRR},
  volume       = {abs/2505.21333},
  year         = {2025},
}

@inproceedings{slake,
  author       = {Bo Liu and
                  Li{-}Ming Zhan and
                  Li Xu and
                  Lin Ma and
                  Yan Yang and
                  Xiao{-}Ming Wu},
  title        = {Slake: {A} Semantically-Labeled Knowledge-Enhanced Dataset For Medical
                  Visual Question Answering},
  booktitle    = {18th {IEEE} International Symposium on Biomedical Imaging, {ISBI}
                  2021, Nice, France, April 13-16, 2021},
  pages        = {1650--1654},
  publisher    = {{IEEE}},
  year         = {2021},
}

@article{pmc,
  author       = {Xiaoman Zhang and
                  Chaoyi Wu and
                  Ziheng Zhao and
                  Weixiong Lin and
                  Ya Zhang and
                  Yanfeng Wang and
                  Weidi Xie},
  title        = {{PMC-VQA:} Visual Instruction Tuning for Medical Visual Question Answering},
  journal      = {CoRR},
  volume       = {abs/2305.10415},
  year         = {2023},
}

@inproceedings{medxpertqa,
  author       = {Yuxin Zuo and
                  Shang Qu and
                  Yifei Li and
                  Zhang{-}Ren Chen and
                  Xuekai Zhu and
                  Ermo Hua and
                  Kaiyan Zhang and
                  Ning Ding and
                  Bowen Zhou},
  editor       = {Aarti Singh and
                  Maryam Fazel and
                  Daniel Hsu and
                  Simon Lacoste{-}Julien and
                  Felix Berkenkamp and
                  Tegan Maharaj and
                  Kiri Wagstaff and
                  Jerry Zhu},
  title        = {MedXpertQA: Benchmarking Expert-Level Medical Reasoning and Understanding},
  booktitle    = {ICML},
  series       = {Proceedings of Machine Learning Research},
  publisher    = {{PMLR} / OpenReview.net},
  year         = {2025}
}
